\documentclass{ieeeaccess}
\usepackage{placeins}
\renewcommand{\thetable}{\Roman{table}}
\usepackage{cite}
\usepackage{amsmath,amssymb,amsfonts}
\usepackage{graphicx}
\usepackage{textcomp}
\usepackage{tabulary}
\usepackage{algorithm}%
\usepackage{algpseudocode}%
\usepackage[caption=false,font=footnotesize]{subfig}
\usepackage{multirow}
\usepackage{rotating}  
\usepackage{bm}
\usepackage{booktabs}
\usepackage{url}
\usepackage{needspace}
\usepackage{float}

\usepackage{bm}
\makeatletter
\AtBeginDocument{\DeclareMathVersion{bold}
\SetSymbolFont{operators}{bold}{T1}{times}{b}{n}
\SetSymbolFont{NewLetters}{bold}{T1}{times}{b}{it}
\SetMathAlphabet{\mathrm}{bold}{T1}{times}{b}{n}
\SetMathAlphabet{\mathit}{bold}{T1}{times}{b}{it}
\SetMathAlphabet{\mathbf}{bold}{T1}{times}{b}{n}
\SetMathAlphabet{\mathtt}{bold}{OT1}{pcr}{b}{n}
\SetSymbolFont{symbols}{bold}{OMS}{cmsy}{b}{n}
\renewcommand\boldmath{\@nomath\boldmath\mathversion{bold}}}
\makeatother

\def\BibTeX{{\rm B\kern-.05em{\sc i\kern-.025em b}\kern-.08em
    T\kern-.1667em\lower.7ex\hbox{E}\kern-.125emX}}

\makeatletter

\let\IEEEoriginaltitlepage\ps@titlepage
\def\ps@titlepage{%
    \IEEEoriginaltitlepage
    \def\@oddhead{\rule{0pt}{14pt}}%
    \def\@evenhead{\rule{0pt}{14pt}}%
}

\let\IEEEoriginalheadings\ps@headings
\def\ps@headings{%
    \IEEEoriginalheadings
    \def\@oddhead{\rule{0pt}{14pt}}%
    \def\@evenhead{\rule{0pt}{14pt}}%
}

\renewcommand{\historyfont}{%
    \sffamily
    \fontencoding{T1}
    \fontfamily{formata}
    \fontseries{n}
    \fontshape{n}
    \fontsize{7}{8.4}
    \selectfont
    \color{white}%
}

\renewcommand{\doifont}{%
    \rmfamily
    \fontencoding{T1}
    \fontfamily{giovannistd}
    \fontseries{n}
    \fontshape{it}
    \fontsize{6}{7.2}
    \selectfont
    \color{white}%
}

\makeatother

\AtBeginDocument{\pagestyle{headings}}

\makeatletter

\renewcommand{\footervolfont}{%
    \sffamily
    \fontencoding{T1}
    \fontfamily{formata}
    \fontseries{n}
    \fontsize{6}{7}
    \selectfont
    \color{white}%
}

\makeatother

\begin{document}
\history{}
\doi{}

\title{Deep Learning-Based Classification of Cognitive and Resting States Using Electroencephalography Signals}
\author{\uppercase{K. A. Januka S. Fernando}\authorrefmark{1}, and
\uppercase{Harshit Srivastava}\authorrefmark{2}}

\address[1]{School of Computing Technologies, RMIT University, 124 La Trobe Street, Melbourne, VIC 3000, (e-mail: s4151141@student.rmit.edu.au)}
\address[2]{Department of CSE (AI) and CSE (AI \& ML), Institute of Engineering \& Management, Kolkata, University of Engineering and Management, Kolkata - 700156 (e-mail: harshit.srivastava@uem.edu.in)}

\markboth
{K. A. J. S. F. \headeretal: Preparation of Papers for IEEE TRANSACTIONS and JOURNALS}
{K. A. J. S. F.  \headeretal: Preparation of Papers for IEEE TRANSACTIONS and JOURNALS}

\corresp{Corresponding author: K. A. Januka S. Fernando (e-mail: s4151141@student.rmit.edu.au).}

\begin{abstract}
The categorization of cognitive and resting states derived from electroencephalography (EEG) signals is crucial for comprehending fluctuations in brain activity linked to various mental states. EEG provides a non-intrusive approach for documenting brain function in both resting and task-oriented cognitive conditions, whilst deep learning techniques enable the automatic extraction of significant patterns from intricate EEG data. This study presents a deep learning framework to distinguish between resting and cognitive states through EEG records. The proposed framework integrates a Convolutional Neural Network (CNN) stacked with a Gated Recurrent Unit (GRU) for the extraction of features from EEG signals. Time-frequency analysis is conducted to explore the salient aspects of signals, and the derived features are then assessed utilizing conventional deep learning and machine learning classifiers, including the suggested 2D-Net architecture. The proposed approach and feature extraction strategy outperform the evaluated comparative methods, achieving accuracies of 83.177\% for resting-versus-mathematical task classification, 76.107\% for resting-versus-memory task classification, and 83.432\% for resting-versus-music task classification. The findings illustrate the efficacy of integrating signal processing with deep learning methodologies to discriminate resting from cognitive states utilizing EEG signals.
\end{abstract}


\titlepgskip=-21pt

\maketitle 

\section{Introduction}
\label{sec:introduction}
\PARstart{E}{lectroencephalography} (EEG) is a method for capturing the brain's electrical activity through a non-invasive platform, which is extensively employed to investigate fluctuations linked to various mental conditions. By analyzing EEG signals, classification methods can be used to distinguish patterns associated with cognitive activity and resting conditions. EEG thus serves as a significant method for examining cerebral activity during both task performance and periods of rest. \cite{b1, b2}. 

Signal processing plays an important role in EEG analysis by reducing noise and enhancing signal components relevant to subsequent interpretation and classification. Bandpass filtering is commonly applied to suppress unwanted frequency components while preserving frequency ranges relevant to the analysis. Prior research has examined several filtering methodologies \cite{b3, b4, b5} and utilized bandpass filtering as a preliminary measure in EEG-related activities including epileptic seizure identification and sleep-stage assessment \cite{b6, b7, b8}. Independent Component Analysis (ICA) is a widely utilized technique that separates multichannel EEG recordings into statistically independent components, facilitating the detection and elimination of artifacts and other undesirable signal sources. Previous studies have applied ICA to EEG classification \cite{b9} and artifact removal \cite{b10, b11}.

The study in the frequency domain yields further data regarding the spectrum features of EEG signals, highlighting patterns related to frequently examined brain rhythms including delta, theta, alpha, and beta activity. The Fourier Transform and its computer counterpart, the Fast Fourier Transform (FFT), are extensively utilized to depict EEG signals in the frequency domain and have been employed in applications such as emotion recognition and the analysis of epileptic EEGs. \cite{b12, b13}. Time-frequency analysis extends this perspective by examining how spectral characteristics evolve over time. Methods like the Wavelet Transform and Short-Time Fourier Transform (STFT) can detect fleeting alterations and temporal fluctuations in EEG activity that might remain obscured when examining the time or frequency domain solely. Previous studies have used time-frequency features for applications such as music preference recognition, epileptic seizure detection, and motor imagery classification \cite{b14, b15, b16}. Together, time, frequency, and time-frequency domain representations provide complementary perspectives for EEG analysis and can support the extraction of informative patterns for classification.

\subsection{Problem definition}
Identifying differences between resting and cognitively active states from EEG recordings remains an important problem in the analysis of brain activity. The present study investigates whether EEG signals contain sufficiently informative patterns to distinguish an eyes-open resting condition from task-related cognitive activity. In particular, the objective is to determine whether variations in EEG signals can be used to differentiate a resting state from cognitive conditions associated with mathematical processing, memory recall, and music-related mental activity. Although previous research has investigated eye-state classification and the recognition of individual mental tasks, distinguishing resting activity from different cognitive conditions remains a relevant challenge for EEG-based classification. This research formulates and assesses a signal-processing and deep learning framework that combines feature extraction with classification methods to distinguish the eyes-open resting state from the three cognitive tasks.

\subsection{Motivation}
EEG offers a significant method for analyzing alterations in cerebral activity linked to various cognitive states. Distinguishing resting and cognitively active states can help characterize variations in EEG patterns and improve the understanding of how neural activity changes during different mental processes. These contrasts are pertinent to EEG-based applications, such as brain-computer interfaces and neurofeedback systems, where insights into a person's cognitive state can enhance system reactions or feedback.

The cognitive tasks considered in this study represent different forms of mental activity, including mathematical processing, memory-related activity, and music-related cognition. Investigating whether these tasks can be distinguished from an eyes-open resting condition provides a useful setting for examining task-related changes in EEG signals. In particular, mathematical and memory tasks involve cognitive processes that can be reflected in changes in neural activity, while music-related activity provides an additional cognitive condition with distinct task characteristics. Therefore, the classification of these states offers an opportunity to investigate whether informative EEG representations can distinguish resting activity from multiple forms of cognitive engagement.

The subsequent sections of this document are structured as outlined below. Section \ref{Sec:2} offers the foundational context and delineates the methodologies pertinent to this research. Section \ref{Sec:3} reviews related work and discusses the findings and limitations of existing studies. \ref{Sec:4} presents the materials and methodology, including the data collection protocol, experimental procedures, and performance metrics. The experimental results are discussed in \ref{Sec:5}. Ultimately, sections \ref{Sec:6} and \ref{Sec:7} provide the conclusion and outline directions for future research.

\section{Background study}
\label{Sec:2}
\begin{figure}[h]
\centering
\includegraphics[width=7cm]{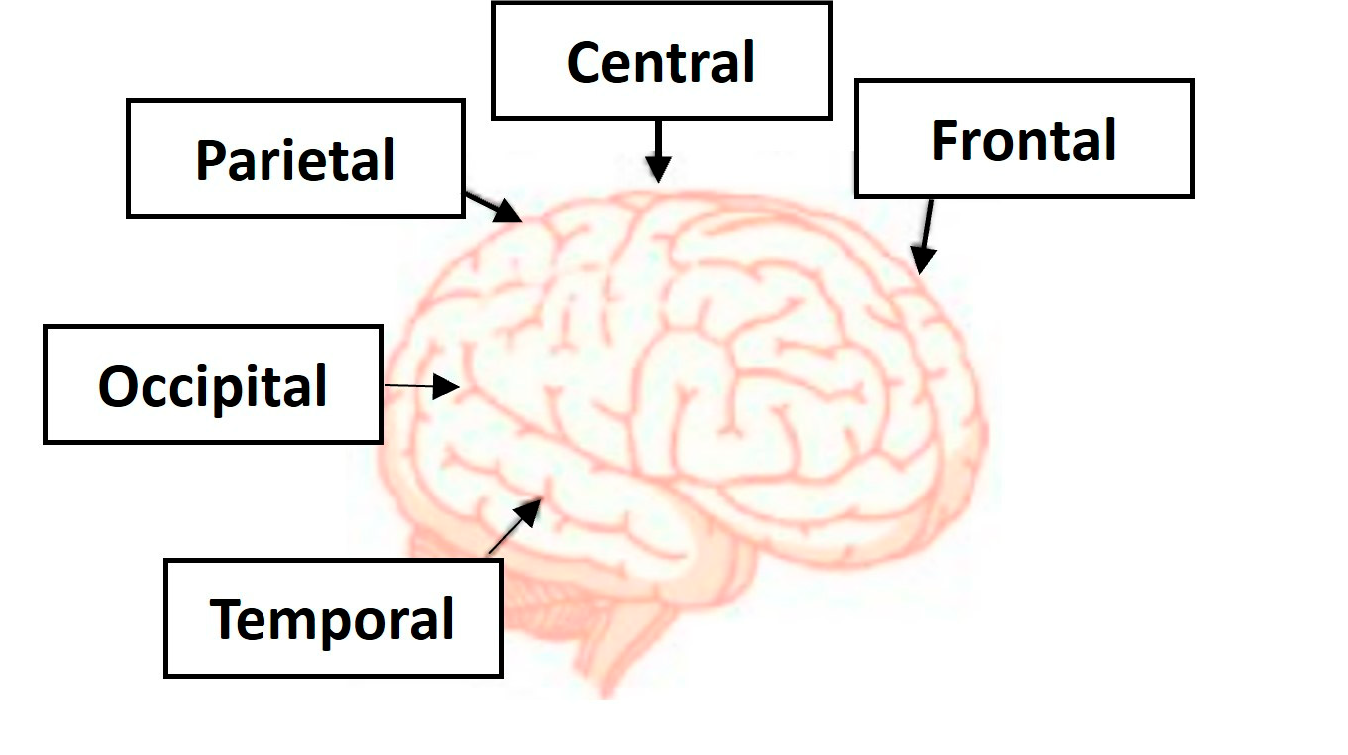}
\caption{Different lobes in the human brain}
\label{eegbrain}
\end{figure}

Hans Berger, the first to record electrical signals obtained from the human brain (1924), reported his findings four years later. Berger, a psychiatrist at the University of Jena in Germany, played a foundational role in establishing EEG as a method for recording human brain activity \cite{b17}. Fig. \ref{eegbrain} illustrates that the human brain is mainly divided into four lobes: the frontal lobe, parietal lobe, temporal lobe, and occipital lobe. The central region is situated between the frontal lobe and parietal lobe.

\begin{figure}[h]
\centering
\includegraphics[width=7cm]{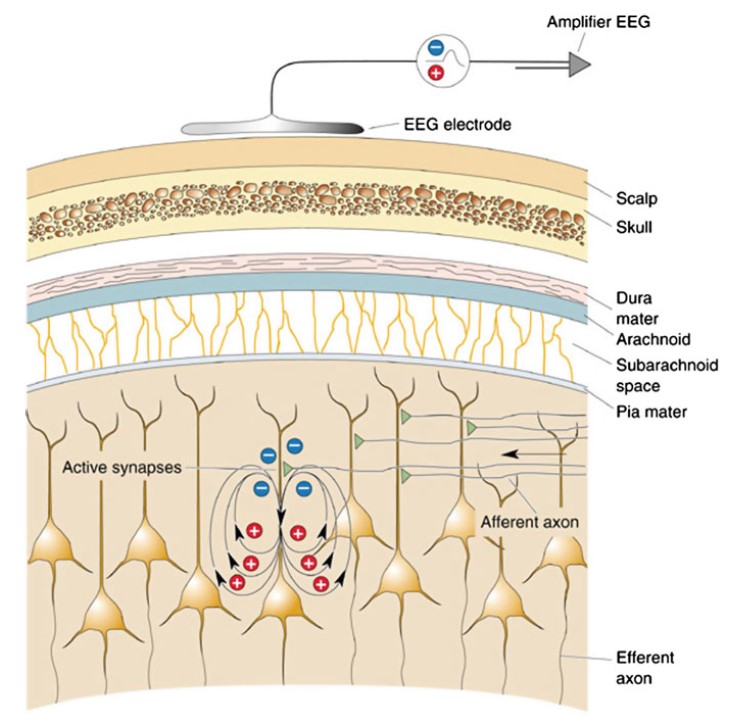}
\caption{Generation of very small electrical fields and electrode measurement}
\label{eeg1}
\end{figure}

EEG recordings are obtained using small disc-shaped electrodes that are temporarily positioned at specific locations on the scalp. The brain's electrical activity is captured by an EEG recording system and an amplifier that are connected to the electrodes. Because these electrical signals are extremely small, they are amplified before being displayed as waveform patterns or stored digitally for further examination. The recorded EEG signals can then be interpreted to examine variations in brain activity, as illustrated in Fig. \ref{eeg1}.

The EEG electrodes placement on the scalp is consistent with standardized positioning systems, such as the International 10–20 system. In this system, the brain region associated with an electrode position is denoted by letter labels. The occipital, parietal, central, temporal, and frontal regions are represented by O, P, C, T, and F, respectively. The suffix "z" is used to indicate the location of electrodes along the midline, while odd and even numbers indicate the left and right hemispheres, respectively. EEG recordings represent voltage differences measured between electrodes, and the recorded signals can be displayed using different configurations or montages that define the electrode arrangement \cite{b20}.

EEG is significantly used in the investigation and assessment of a range of behavioral and neurological conditions, including epilepsy, sleep disorders, dementia, head injuries, attention-related difficulties, and developmental conditions. In addition to examining the recorded waveform directly, the frequency composition of EEG data may be assessed to extract insights on the brain activity patterns linked to several physiological and cognitive states. Several frequency bands, such as delta, theta, alpha, beta, and gamma, are frequently employed to describe EEG activity. It is possible to analyze the variations associated with various mental and physiological states using these frequency components, which serve as a valuable representation of EEG activity \cite{b21}.

Digital signal processing techniques can be employed during preprocessing to improve the quality of EEG signals and prepare them for subsequent analysis, prior to their analysis and classification. The following subsections describe the key techniques considered in this study and their respective roles in processing and extracting informative characteristics from the EEG signals.

\subsection{Independent Component Analysis (ICA)}
Independent Component Analysis (ICA) functions as a signal processing technique that separates multichannel EEG data into a collection of statistically independent components. In EEG analysis, these components can be examined through their spatial distributions and temporal activity to distinguish neural patterns from components associated with noise or artifacts. This makes ICA useful for separating potentially informative brain activity from unwanted signal contributions and for selecting components relevant to subsequent analysis \cite{b9, b10, b11}.

FastICA is one of the commonly used algorithms for performing ICA. It estimates a linear transformation that separates the observed EEG recordings into components with as much statistical independence as possible. The foundational concept posits that the detected signals are mixtures of unidentified source signals, and the method seeks to establish a demixing matrix capable of retrieving these hidden sources \cite{b18}.

\subsection{Short-Time Fourier Transformation (STFT)}
The Short-Time Fourier Transform (STFT) offers a time-frequency depiction of a signal by analyzing its spectral characteristics over confined time segments \cite{b19}. In this methodology, the EEG signal is partitioned into overlapping segments, and the Fourier transform is executed on each segment via a window function. The resultant representations illustrate the temporal variations of the signal's frequency components. The Short-Time Fourier Transform is articulated as:

\begin{equation}
STFT(x(t), \tau, \omega) = \int_{-\infty}^{\infty} x(t) w(t - \tau) e^{-i\omega t} dt
\label{eq1}
\end{equation}

where \(x(t)\) denotes the EEG signal in the time domain, \(w(t-\tau)\) represents the window function centered at time \(\tau\), and \(\omega\) denotes the frequency variable. The parameter \(\tau\) determines the temporal position of the analysis window.

The Short-Time Fourier Transform (STFT) utilizes the Fourier transform on consecutive overlapping segments to monitor temporal variations in the frequency attributes of EEG data. This time-localized depiction is advantageous for investigating temporal fluctuations and ephemeral frequency trends that might not be sufficiently captured by traditional frequency-domain research. In this research, a Hann window is utilized prior to the Fourier transform to mitigate spectral leakage. The Hann window is characterized as follows:

\begin{equation}
w(n) = 0.5 \left(1 - \cos\left(\frac{2\pi n}{N-1}\right)\right)
\label{eq2} 
\end{equation}

where \(w(n)\) denotes the window value at sample index \(n\), and \(N\) represents the window length \cite{b14, b15, b16}. \\


\subsection{Gated Recurrent Unit (GRU)}
\begin{figure}[h]
\begin{center}
\includegraphics[width=\linewidth]{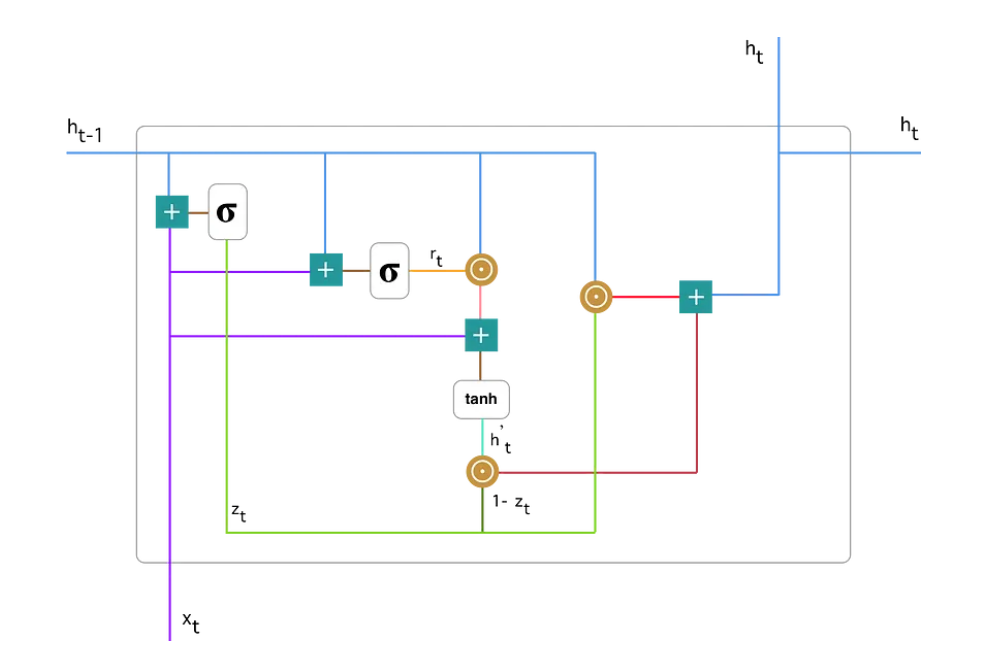}
\caption{GRU - Workflow diagram}
\label{gru}
\end{center}
\end{figure}
\newcommand{\dd}[1]{\mathrm{d}#1}

The Gated Recurrent Unit (GRU) is an architecture of recurrent neural networks intended to capture sequential information while mitigating issues related to learning long-term relationships in standard RNNs. A GRU utilizes two essential gating mechanisms: the reset and the update gates. The reset gate regulates the former concealed state's impact while integrating incoming input, whereas the update gate specifies the extent to which the prior state is maintained in developing the new representation \cite{b22}.

The reset and update gates are computed as

\begin{equation}
r_t = \sigma(W_r x_t + U_r h_{t-1} + b_r)
\label{eq3}
\end{equation}
\begin{equation}
z_t = \sigma(W_z x_t + U_z h_{t-1} + b_z)
\label{eq4}
\end{equation}

In this scenario, \(r_t\) indicates the reset gate while \(z_t\) indicates the update gate. \(x_t\) represents the input at time \(t\), and \(h_{t-1}\) is the hidden state from the last time step.

The fresh hidden state \(h_t\) is generated by combining the former hidden state with the candidate activation using the update gate:

\begin{equation}
h_t = (1 - z_t) \odot h_{t-1} + z_t \odot \tilde{h_t}   
\label{eq5}
\end{equation}

where \(\odot\) represents the operation of multiplying corresponding elements. The update gate \(z_t\) dictates the comparative impact of the preceding concealed state and the potential activation. If \(z_t=0\), the former hidden state is maintained; conversely, \(z_t=1\) leads to the direct application of the candidate activation \(\tilde{h_t}\) as the new hidden state. The reset gate \(r_t\) delineates to what extent the information from the prior concealed state affects the determination of the candidate activation.

Overall, the GRU controls information propagation across successive time steps through its reset and update gates, enabling relevant temporal information to be retained while reducing the influence of less useful information.

\subsection{1D Convolutional Neural Network (1D-CNN)}

A one-dimensional Convolutional Neural Network (1D-CNN) is designed to learn local patterns from sequential data using one-dimensional convolutional filters. This makes it suitable for extracting informative features from time-series signals such as EEG recordings. The input for a 1D-CNN is depicted as a series of data points, with convolutional filters traversing this series to execute element-wise multiplication and subsequent summation \cite{b23}.

The convolution procedure generates feature maps that illustrate the activation of each filter at various locations in the input sequence. These activations encapsulate regional trends within the signal. A max-pooling layer may then be employed to reduce the dimensionality of the feature maps while retaining the most prominent activations in each pooling area. This process offers a succinct depiction of the derived features and diminishes the computational intricacy of following layers.

The combined feature maps are subsequently transformed into a one-dimensional vector suitable for input into fully linked or other following layers for classification. In this way, 1D-CNNs can learn hierarchical representations from sequential data by progressively extracting local features while reducing the dimensionality of the intermediate representations. The operation of a 1D convolutional layer is expressed as

\begin{equation}
y_i = \sum_{j=0}^{k-1} w_j \cdot x_{i+j} + b
\label{eq6}
\end{equation}

where $y_i$ denotes the output value at position $i$, $w_j$ represents the convolutional weights, $x_{i+j}$ denotes the corresponding input values, $b$ is the bias term, and $k$ represents the kernel size.

\section{Related work}
\label{Sec:3}

Research on distinguishing resting and cognitively active states from EEG signals has expanded as researchers seek to characterize changes in brain activity and improve EEG-based systems. Recent research has progressively utilized machine learning and deep learning techniques on EEG signals to identify cognitive, resting, and task-related mental states. These approaches have explored different signal representations, feature extraction strategies, and classification architectures to improve the automatic identification of patterns associated with distinct mental conditions \cite{b24}.

Anderson \emph{et al.} \cite{b25} investigated five cognitive tasks using recordings from four participants. Their methodology utilized feed-forward neural networks including two or three layers and was assessed by 10-fold cross-validation. The main limitation was the small participant pool, together with the restricted comparison of classification methods. The analysis also focused only on artificial neural networks.

Wang \emph{et al.} \cite{b26} presented a test-retest EEG dataset that includes both eyes-open and eyes-closed resting states, as well as three participant-induced cognitive conditions: memory recall, mental singing, and serial subtraction. The dataset was collected from 60 participants across three recording sessions and was designed to support the investigation of both within-session and between-session variability in EEG measures. This dataset provides the basis for the experiments conducted in the present study.

Bai \emph{et al.} \cite{b27} examined the classification of mental tasks through EEG signals utilizing several deep learning frameworks, such as convolutional neural networks (CNN), long short-term memory (LSTM), and, gated recurrent unit (GRU) architectures. The study also proposed a hybrid architecture that combined recurrent processing with a CNN decoder. Their results demonstrated the potential of combining convolutional and recurrent architectures for classifying mental tasks based on EEG data.

Craik \emph{et al.} \cite{b24} conducted a review on the application of deep learning techniques for the EEG signal classification in various domains. These encompassed motor imaging, emotion identification, cognitive load assessment, seizure identification, event-related potentials, and sleep stage evaluation. Siddiqui \emph{et al.} \cite{b28} created a deep neural network for classifying mental tasks independently of the subject, utilizing a standard EEG dataset. Their approach included power spectral density analysis and principal component analysis (PCA) for the extraction of features and dimensionality reduction. The study also noted that part of the available data had to be retained for testing, which restricted the data available for training the model.

Lee \emph{et al.} \cite{b29} investigated continual learning for motor imagery-based BCI using EEG recordings collected from five participants. Their methodology utilized a transformer-driven spatial-temporal network to interpret motor imagery intentions over consecutive training sessions. The study demonstrated improved classification performance through continual learning, but the authors identified the small sample size and computational requirements of the transformer-based framework as important limitations.

Qayyum \emph{et al.} \cite{b30} used a 1D-CNN to distinguish resting and cognitive states while estimating cognitive load from alpha-band activity. Discrete Wavelet Transform (DWT) was used to derive brainwave features, and the 1D-CNN produced consistent results for cognitive-load estimation.

Ahmad \emph{et al.} \cite{b1} also examined the separation of cognitive and resting conditions using both linear and nonlinear features. Their experimental protocol included a five-minute eyes-open resting period succeeded by an IQ task based on Raven's Advanced Progressive Matrix (RAPM). Wavelet-based features and sample entropy were extracted, and an SVM was used for classification.

Mazher \emph{et al.} \cite{b31} examined the categorization of resting and cognitive states through EEG recordings obtained during both resting and learning scenarios. Their study involved 34 participants and extracted frequency-band information using the Discrete Wavelet Transform (DWT), together with spectral and approximate entropy features. Effective connectivity measures were also evaluated using Partial Directed Coherence (PDC), with an Extreme Learning Machine (ELM) used for classification. The highest classification accuracy obtained from the connectivity-based approach was 79.90\% for the alpha band.

Tigga \emph{et al.} \cite{b32} introduced the AttGRUT model for detecting EEG abnormalities associated with depression. The feature set included statistical, spectral, and wavelet-based measures. Recursive feature elimination and the Boruta method were then used for feature selection, with Shapley-based explanations included to interpret the selected features.

Liang \emph{et al.} \cite{b33} evaluated an Extreme Learning Machine for classifying five mental tasks from EEG recordings and compared its performance with BPNN and SVM models. Their results considered both classification accuracy and training time and showed that refining the classifier outputs could improve the final performance.

Li \emph{et al.} \cite{b34} investigated EEG-based mental task identification using wavelet packet entropy and an SVM classifier. A db4 wavelet was used for the wavelet decomposition, and the findings demonstrated the usefulness of wavelet-based entropy features for distinguishing mental tasks.

Recently, Panwar \emph{et al.} \cite{b35} examined cognitive state evaluation by EEG brain connectivity and a deep learning model. Their study considered multiple cognitive tasks associated with workload, attention, and fatigue, demonstrating the continued use of evaluation of cognitive states. However, the approach relied on connectivity-based representations, which differ from the time-frequency representation used in the present study.

Wang \emph{et al.} \cite{b36} introduced LGNet for the categorization of cognitive workload derived on EEG data through the integration of local and global representations. The architecture employed convolutional techniques to seize localized EEG features and supplementary methods to represent wider interdependencies. The research underscores the capability of deep architectures to derive complementary representations from EEG data, simultaneously demonstrating the growing adoption of intricate designs for cognitive-state classification.

Sharma and Gupta recently investigated EEG-based mental workload detection using resting intervals and arithmetic tasks. Their strategy amalgamated Discrete Wavelet Transform (DWT) with Welch's power spectral density estimation before analyzing standard machine learning approaches and a deep learning architecture utilizing LSTM. Their findings further demonstrate the usefulness of combining frequency-domain signal processing with deep learning for distinguishing EEG-based mental workload states \cite{b37}.

Several limitations can be observed in existing studies. Many works rely on relatively small sample sizes, raising concerns about generalizability and scalability. The majority of methods pay little attention to frequency or time-frequency characteristics and instead concentrate mostly on temporal variables. Although both deep learning and conventional machine learning models have been used, only a small number of classifiers have been investigated. In addition, resting and cognitive states are often treated separately, whereas combining them may provide more comprehensive insights into brain activity.\\

The key contributions of this study are highlighted below:

\begin{itemize}
  \item EEG recordings from 40 participants are analyzed to evaluate the ability of the proposed framework to distinguish resting and cognitive states across multiple subjects.
  
  \item The proposed analysis considers both frequency-domain and time-frequency information, allowing the characteristics of EEG signals to be examined from complementary perspectives.
  
  \item A combined signal-processing and deep learning pipeline is developed in which bandpass filtering, Independent Component Analysis (ICA), and Short-Time Fourier Transform (STFT) are used with a CNN-GRU-based feature extraction approach. This combination is used to capture informative patterns from the EEG recordings for subsequent classification.
  
  \item A 2D-CNN architecture, referred to as 2D-Net, is introduced for three binary classification tasks: EO-MA, EO-ME, and EO-MU. To deliver a more comprehensive assessment of classification performance, its performance is evaluated by precision, recall, F1-score, and Cohen's kappa.
\end{itemize}

The classification tasks considered in this study involve distinguishing the eyes-open resting condition from three forms of cognitive activity: continuous mathematical subtraction, memory recall of daily activities, and mentally singing a preferred song. These task combinations provide a specific setting for examining differences between resting and task-related EEG activity. The proposed feature extraction pipeline and 2D-Net classifier are therefore evaluated in this context to investigate their effectiveness in distinguishing these states.

\section{Materials and methodology}
\label{Sec:4}

This portion outlines the dataset, data preparation protocols, analysis techniques, and recommended methodologies utilized in the investigation.

\subsection{Data description}

\begin{figure}[h]
\begin{center}
\includegraphics[width=7cm]{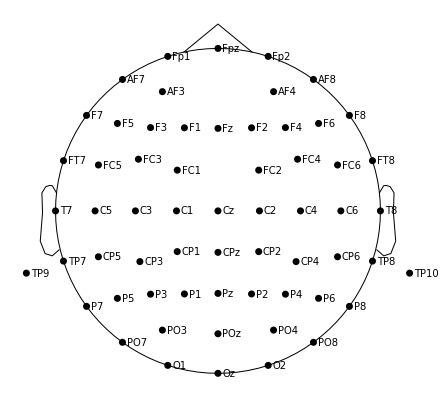}
\caption{62 Channel placements}
\label{fig0}
\end{center}
\end{figure}

The publicly available ``Test-Retest Resting, and Cognitive State EEG Dataset'' introduced by Y. Wang \emph{et al.} \cite{b40} was used for the experiments. The dataset contains recordings from 60 undergraduate students within a relatively narrow age range. The average age of the participants was 20.01 years, including 32 girls and 28 men. EEG signals were obtained through the use of 63 or 64 Ag/AgCl active electrodes placed following the enlarged international 10-20 technique. The recording reference was established using the FCz electrode, with signals captured at 500 Hz. To maintain a consistent channel configuration across the different electrode caps, the recordings were standardized to 62 channels, as shown in Fig. \ref{fig0}.

{
\linespread{1}
\begin{table*}[h]
\centering
\caption{Cognitive states obtained through the experiment}
\begin{tabular}{ll}
\toprule
\textbf{Task} & \textbf{Description} \\
\midrule
The tasks involved memory & Subjects were asked to recollect their day's activities\\
\hline
The tasks involved music & Subjects were instructed to mentally sing their favorite songs. \\
\hline
The tasks involved subtraction & Subjects were instructed to count backward by 7s from 5000.  \\
\bottomrule
\end{tabular}
\end{table*}
}

Each participant completed three trials. The second trial was performed 90 minutes after the first trial, while the third was conducted one month later as part of the retest protocol. In each trial, subjects engaged in 5 minutes of resting with eyes open, followed by 5 minutes of resting with eyes closed, then performed three cognitive activities, each lasting 5 minutes.

For the classification experiments, 40 participants were selected to minimize the computational complexity of the training process, utilizing just the initial trial from each individual. The eyes-closed (EC) condition was excluded because it was outside the scope of the present analysis. The remaining conditions were therefore used for the classification experiments.

\subsection{Data preprocessing}

The EEG recordings were collected and preprocessed by the authors of \cite{b40}. Each experimental condition provided 5 minutes of EEG data for every participant. The raw recordings were processed in EEGLAB using a sequence of procedures aimed at minimizing noise and improving signal quality for subsequent analysis.

Initially, the records were adjusted to a standard average reference. A symmetric finite impulse response (FIR) filter was subsequently utilized, preserving frequencies ranging from 0.3 to 45 Hz. The resulting signals were visually examined to identify channels with substantial recording problems. A channel was classified as problematic when at least one-third of its trials were affected. No channels met this criterion in the dataset.

Missing channel information was subsequently addressed using linear interpolation, in which the missing values were estimated from neighboring channels. After interpolation, the recordings were again re-referenced using the common average reference. The persistent EEG recordings were later partitioned into 4-second epochs. Each epoch was scrutinized by hand, and those with inadequate signal quality were discarded. No flawed epochs were recognized within the dataset.

\begin{figure}[h]
\begin{center}
\includegraphics[width=8cm]{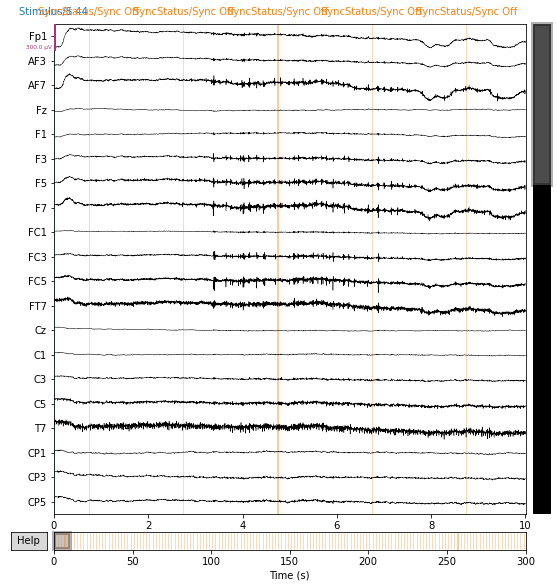}
\caption{Signal representation across channels after preprocessing of a single subject}
\label{fig12}
\end{center}
\end{figure}

For the eyes-open recordings, independent components associated with artifacts, including eye blinks and eye movements, were identified and marked during the preprocessing procedure. The resulting preprocessed EEG signals for all channels across the full 300-second recording period of a representative participant are shown in Fig. \ref{fig12}.

\subsection{Frequency analysis}

Frequency analysis was conducted prior to categorization to investigate the spectral properties of various mental states and pinpoint frequency ranges exhibiting significant disparities. The EEG recordings were converted from the time domain to the frequency domain by the Fourier transform \cite{b12, b13}. Welch's approach was employed for the estimation of the power spectral density (PSD) \cite{b38}. In this method, the signal is partitioned into overlapping parts, a periodogram is computed for each segment utilizing the Fast Fourier Transform (FFT), and the averaged periodograms yield the PSD estimate.

Prior to the computation of spectral power, the EEG signals underwent filtration via a Butterworth bandpass filter with a frequency range of 0.5 to 50 Hz. The processed signals were subsequently employed to assess the power within the pertinent frequency ranges. The average power was computed for each band across all EEG channels. These values were subsequently compared across the considered mental states to identify frequency bands with more pronounced differences.

\subsection{Proposed method}

\begin{figure}[h]
\centering
\includegraphics[width=8cm]{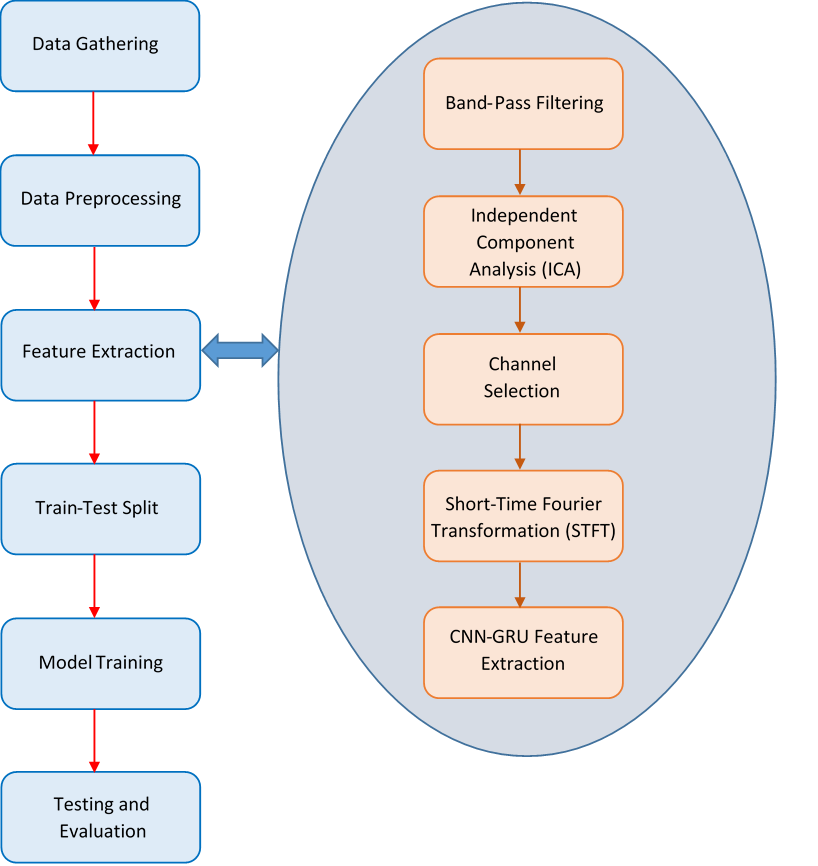}
\caption{Workflow diagram of the complete process}
\label{fig13}
\end{figure}

The suggested framework is segmented into two fundamental elements: feature extraction and classification. The feature extraction stage prepares informative representations from the EEG recordings, which are then used by the classification stage to distinguish the considered states. The complete workflow is illustrated in Fig. \ref{fig13}.

\subsubsection{Feature extraction}

The feature extraction stage combines several signal-processing operations to transform the raw EEG recordings into representations suitable for classification. First, a Butterworth bandpass filter is utilized to retain frequencies between 0.5 and 50 Hz. ICA is then performed on the filtered recordings to segregate the signals into independent components. The squared sum of each component is calculated to determine the most informative channels, and the five channels with the highest values are selected.

The mean amplitude of the selected channels is subsequently calculated to form a representative signal that incorporates information from the chosen channels. STFT is then applied using a Hann window to characterize changes in the signal across time and frequency. The resultant magnitude spectrogram offers a combined time-frequency depiction of the EEG signal.

Ultimately, a CNN-GRU framework is utilized for feature extraction from the time-frequency representation. The CNN component learns informative patterns from the representation, while the GRU captures sequential information. The resulting feature representation is then used for the subsequent classification stage.

\subsubsection{1D CNN-GRU feature extraction}

The 1D CNN-GRU architecture combines convolutional and recurrent processing to obtain informative representations from EEG sequences. The 1D-CNN identifies localized patterns within the input signal, whereas the GRU captures relationships over consecutive time intervals. The output produced by the convolutional stage is supplied to the GRU, allowing the combined network to represent both spatial and temporal characteristics of the EEG data.

\begin{algorithm}[!t]
\caption{1D CNN-GRU-based feature extraction for multi-channel time series}
\label{alg:cnn_gru_feature_extraction_multivariate}
\begin{algorithmic}[1]

\Statex \textbf{Input:} Training/testing sequences $(X,\hat{X})$ and outputs $(Y,\hat{Y})$.
\State Initialize CNN params: filters $f$, kernel size $k$, stride $s$.
\State Initialize GRU params: hidden units $h$, learning rate $\epsilon$, epochs $N$.

\For{$n \gets 1$ \textbf{to} $N$}
  \For{$t \gets 1$ \textbf{to} $\mathrm{len}(X)$}
    \State Apply 1D convolution to $X_t$ using $(f,k,s)$.
    \State Apply max pooling to convolution output.
    \State Feed pooled output to GRU with $h$ hidden units.
  \EndFor
  \State Concatenate final hidden states into feature vector $F_n$.
  \State Compute loss between $\hat{Y}_n$ and $Y_n$.
  \State Update CNN+GRU parameters via backpropagation.
\EndFor

\Statex \textbf{Output:} Extracted feature vector \textit{F} for the multivariate time series \textit{X}.

\end{algorithmic}
\end{algorithm}

The network receives an input with 10 features and uses a sliding window of five consecutive data points. This window provides the model with a short sequence of neighboring observations for learning local relationships. A 1D convolutional layer with 64 filters and a kernel size of 3 is then applied to the input, producing 64 feature maps.

The convolutional outputs are passed through max pooling to reduce their dimensionality while retaining the strongest activations. The pooled representation is subsequently provided to the GRU, which processes the sequence step by step and maintains information from earlier time points. The GRU output is then connected to a dense layer containing 16 nodes with a ReLU activation function. For the binary classification stage, the resulting representation is further reduced from 513 features to 16 through a dense layer. A final single-node layer with sigmoid activation converts this representation into a probability between 0 and 1 corresponding to the two classes.

\subsection{2D-Net - A 2D CNN classifier}

The proposed 2D-Net is a 2D convolutional architecture developed to classify EEG representations in the time-frequency domain. The network interprets the input using a succession of convolutional, normalizing, pooling, and dense layers to learn relevant temporal and spatial patterns.

\begin{figure}[!t]
\begin{center}
\includegraphics[width=3.5cm, height=18cm]{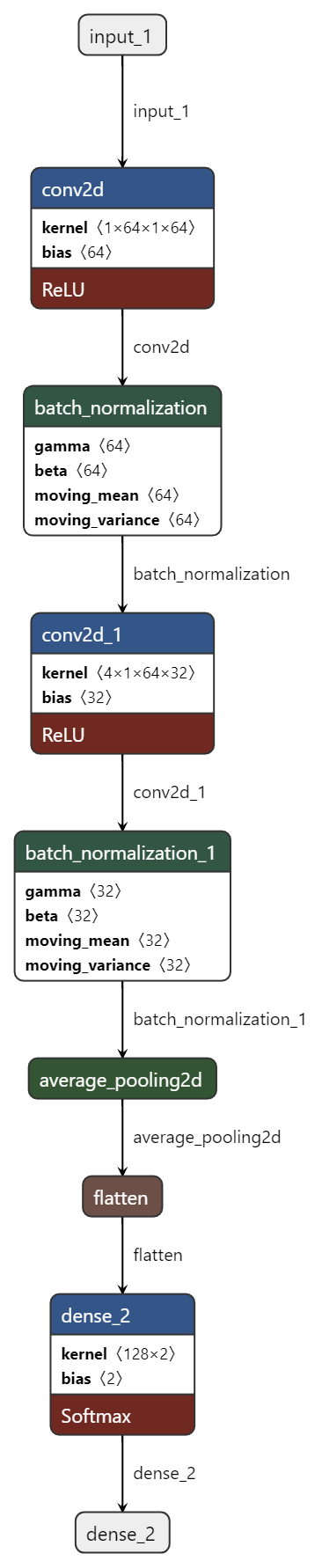}
\caption{Proposed 2D-Net classifier}
\label{final}
\end{center}
\end{figure}

The initial convolutional layer uses filters of size $(1,64)$ to learn local patterns along the temporal dimension. Batch normalization is applied after this layer to support stable training. A second convolutional layer with filters of size $(4,1)$ is then used to capture patterns across the input dimensions, followed by another batch normalization layer. Average pooling utilizing a pool size of $(1,4)$ is then employed to diminish the dimensionality of the feature map while maintaining the essential information for classification. The aggregated feature maps are converted into a one-dimensional format and forwarded to a fully connected output layer. A softmax activation function is employed to produce the anticipated probability for every class. The entire framework of the suggested 2D-Net is depicted in Fig. \ref{final}.

\subsection{Experimental setup}

The processed EEG data were categorized into three binary classification tasks: EO-MA, EO-ME, and EO-MU. Min-max normalization was applied to standardize the input values within a common range and support model training. For each classification task, a participant-wise train-test split was used to prevent data from the same individual from appearing in both sets. Of the 40 participants included in the experiments, 30 participants were assigned to the training set and the remaining 10 participants were reserved for testing, corresponding to a 75:25 split. This separation ensured that the test set consisted entirely of participants unseen during model training, allowing the classification performance to be evaluated under a subject-independent setting. The resulting datasets were then used to train and evaluate the proposed model together with the selected deep learning and machine learning baselines.

The proposed model underwent training for 30 epochs with the Adam optimizer and binary cross-entropy loss. Binary cross-entropy quantifies the disparity between the anticipated probability and the associated target labels, whereas the Adam optimizer adjusts the model parameters throughout training to reduce the loss.

\subsection{Performance metrics}

The classification performance was assessed using accuracy, precision, recall, F1-score, and Cohen's kappa. The corresponding definitions are given below:

\begin{equation}
Accuracy = \frac{TP + TN}{TP + FP + TN + FN}
\label{eq7}
\end{equation}

\begin{equation}
Precision = \frac{TP}{TP + FP}
\label{eq8}
\end{equation}

\begin{equation}
Recall = \frac{TP}{TP + FN}
\label{eq9}
\end{equation}

\begin{equation}
F1-Score = 2 \times \frac{Precision \times Recall}{Precision + Recall}
\label{eq10}
\end{equation}

\begin{equation}
kappa = \frac{P_o - P_e}{1 - P_e}
\label{eq11}
\end{equation}

where TP denotes true positives, TN, FP, and FN denote true negatives, false positives, and false negatives, respectively. In the kappa formulation, $P_o$ represents the observed agreement and $P_e$ represents the expected agreement. Cohen's kappa quantifies the concordance between expected and actual class assignments, while considering the possibility of chance agreement \cite{b39}. Its values fluctuates between $-1$ and $1$, with elevated values signifying greater concurrence.

\section{Experimental results and analysis}
\label{Sec:5}

The experimental assessment was conducted in two phases: frequency and time-frequency domain analysis. The findings derived from these analyzes are detailed in the subsequent subsections.

\subsection{Frequency analysis}

\begin{figure*}[h]
\begin{center}
\includegraphics[width=15cm, height=12cm]{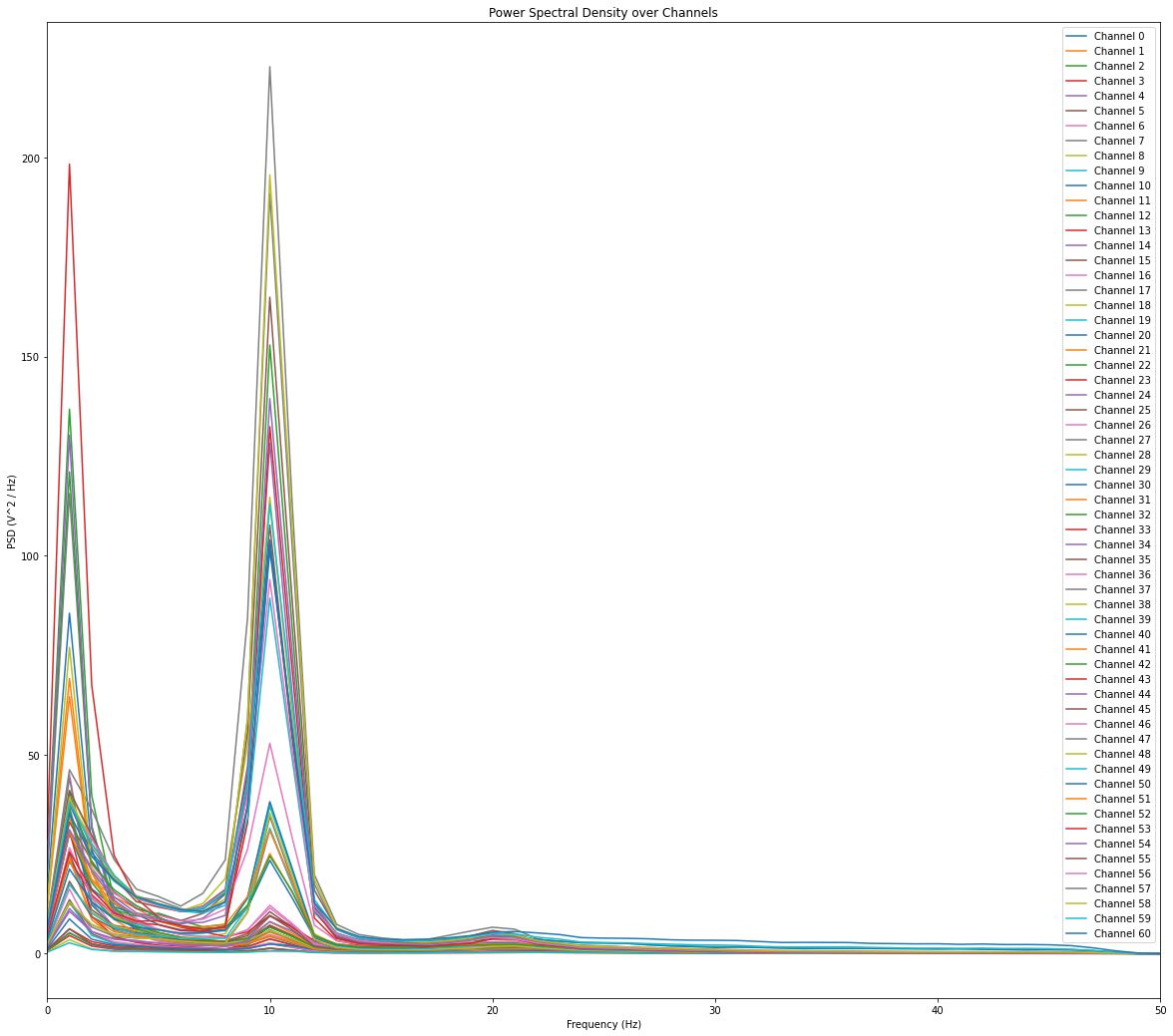}
\caption{PSD over channels - Eyes-closed (EC) state of a single subject}
\label{fig1}
\end{center}
\end{figure*}

\begin{figure*}[!t]
\hspace{0.2cm}
\subfloat[EC - Power of frequency bands]{
    \includegraphics[width=8.6cm,height=5.8cm]{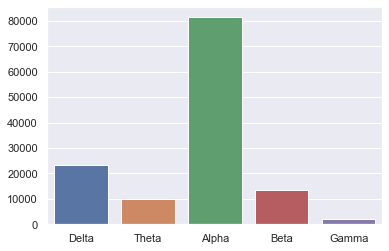}
    \label{fig253a}
}
\hspace{0.3cm}
\subfloat[EC - Power of electrodes at each frequency band]{
    \includegraphics[width=8cm,height=5.8cm]{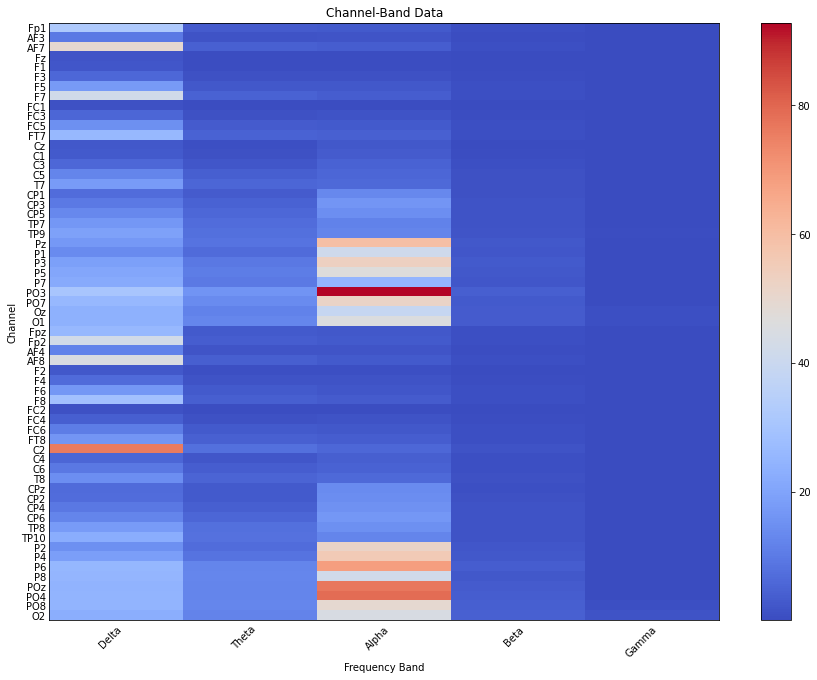}
    \label{fig254a}
}
\caption{Frequency analysis of EC state}
\label{fig2}
\end{figure*}

\begin{figure*}[!t]
\hspace{0.2cm}
\subfloat[EO - Power of frequency bands]{
    \includegraphics[width=8.6cm,height=5.8cm]{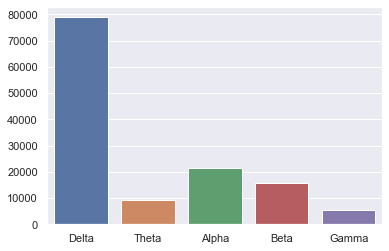}
    \label{fig253b}
}
\hspace{0.3cm}
\subfloat[EO - Power of electrodes at each frequency band]{
    \includegraphics[width=8cm,height=5.8cm]{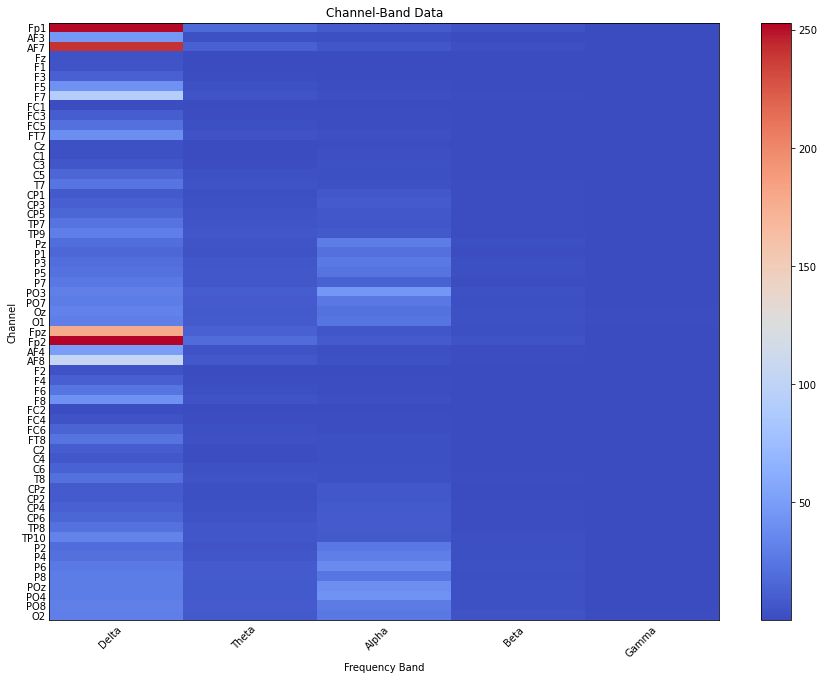}
    \label{fig254b}
}
\caption{Frequency analysis of EO state}
\label{fig4}
\end{figure*}

Figure \ref{fig1} illustrates the power spectral density (PSD) throughout the EEG channels for Subject 4 during the eyes-closed (EC) condition. The EC condition shows stronger activity in the alpha band, with the highest values primarily observed over electrodes located in the parietal region, as illustrated in Fig. \ref{fig2}. In comparison, the eyes-open (EO) condition shows greater activity in the delta band, with stronger responses mainly observed over frontal electrodes, as shown in Fig. \ref{fig4}.

\begin{figure*}[!t]
\hspace{0.2cm}
\subfloat[MA - Power of frequency bands]{
    \includegraphics[width=8.6cm,height=5.8cm]{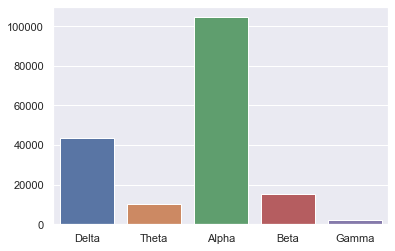}
    \label{fig253c}
}
\hspace{0.3cm}
\subfloat[MA - Power of electrodes at each frequency band]{
    \includegraphics[width=8cm,height=5.8cm]{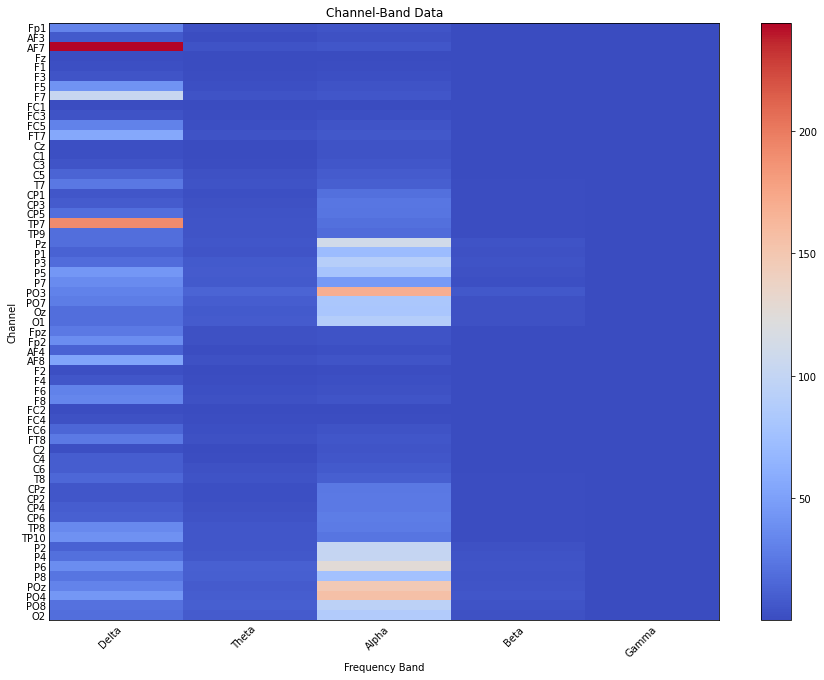}
    \label{fig254c}
}
\caption{Frequency analysis of MA state}
\label{fig7}
\end{figure*}

\begin{figure*}[!t]
\hspace{0.2cm}
\subfloat[ME - Power of frequency bands]{
    \includegraphics[width=8.6cm,height=5.8cm]{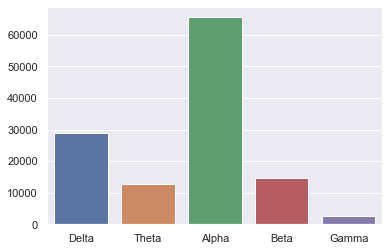}
    \label{fig253d}
}
\hspace{0.3cm}
\subfloat[ME - Power of electrodes at each frequency band]{
    \includegraphics[width=8cm,height=5.8cm]{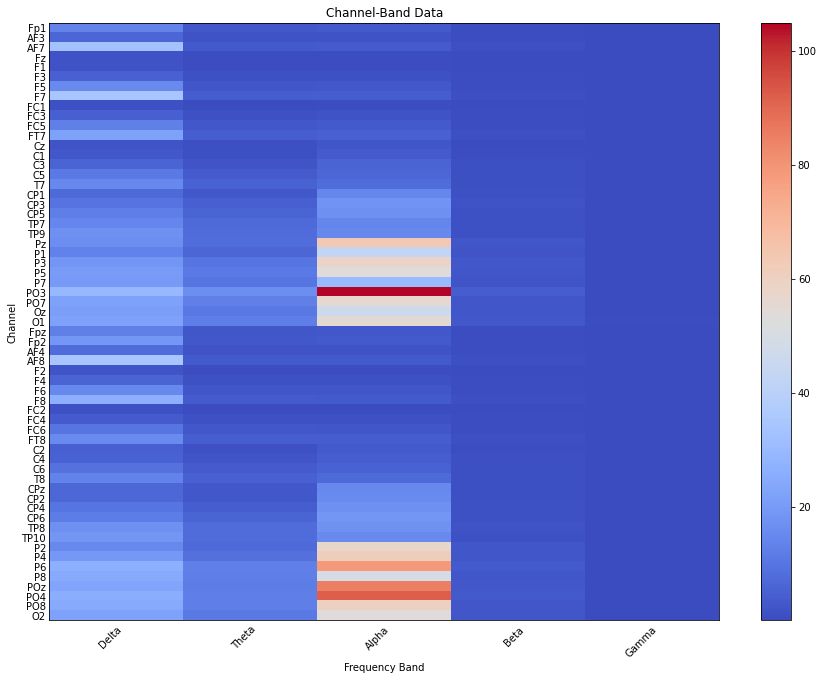}
    \label{fig254d}
}
\caption{Frequency analysis of ME state}
\label{fig9}
\end{figure*}

\begin{figure*}[]
\hspace{0.2cm}
\subfloat[MU - Power of frequency bands]{
    \includegraphics[width=8.6cm,height=5.8cm]{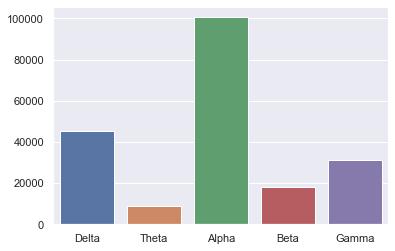}
    \label{fig253e}
}
\hspace{0.3cm}
\subfloat[MU - Power of electrodes at each frequency band]{
    \includegraphics[width=8cm,height=5.8cm]{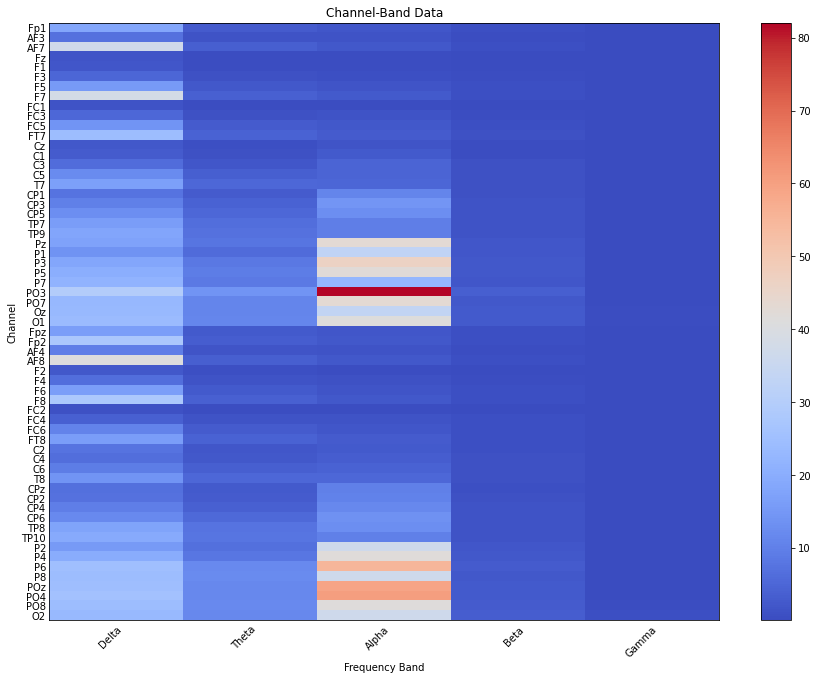}
    \label{fig254e}
}
\caption{Frequency analysis of MU state}
\label{fig11}
\end{figure*}

A relatively prominent alpha-band component is also observed during the subtraction, memory, and music tasks. This activity is mainly concentrated over the parietal electrodes, as shown in Fig. \ref{fig7}, Fig. \ref{fig9}, and Fig. \ref{fig11}.

\subsection{Time-frequency analysis}

Three binary classification tasks were evaluated: EO-MA, EO-ME, and EO-MU, corresponding to the eyes-open resting state and the three cognitive conditions. The evaluation was performed from two perspectives.

First, the proposed 2D-Net was compared with the selected deep learning and machine learning baselines, including ANN, 1D-CNN, LSTM, XGBoost, SVM, and DT. The models were assessed based on accuracy, precision, recall, F1-score, and Cohen's kappa. The 2D-Net was tested both with and without CNN-GRU feature extraction to examine the contribution of the feature extraction stage. Second, different feature extraction approaches were compared using AR, PCA, NNMF, and CNN-GRU. The extracted features were evaluated with the selected machine learning classifiers and the 2D-Net classifier to determine the effect of the feature extraction method on classification performance.

\subsubsection{Resting-Mathematical (EO-MA) states classification}

The EO-MA classification task was evaluated using the selected baseline models and the proposed 2D-Net classifier. The highest accuracy was obtained by 2D-Net with CNN-GRU feature extraction, reaching 83.177\% with a Cohen's kappa of 0.6635. The Decision Tree (DT) produced the next highest accuracy at 82.538\%.

\begin{figure}[!t]
\begin{center}
\includegraphics[width=8.6cm]{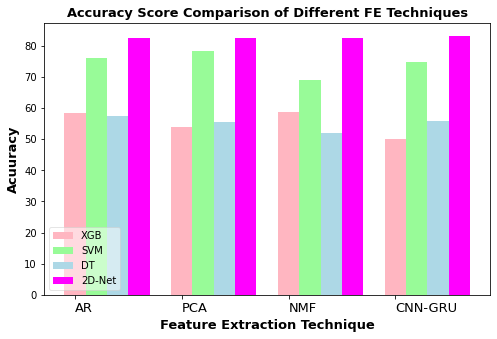}
\caption{Accuracy score (\%) comparison of different feature extraction techniques (EO-MA)}
\label{fig111}
\end{center}
\end{figure}

\begin{figure*}[!t]
\centering
\subfloat[GRU-CNN training vs validation accuracy (\%) (EO-MA)]{
  \includegraphics[width=0.32\textwidth]{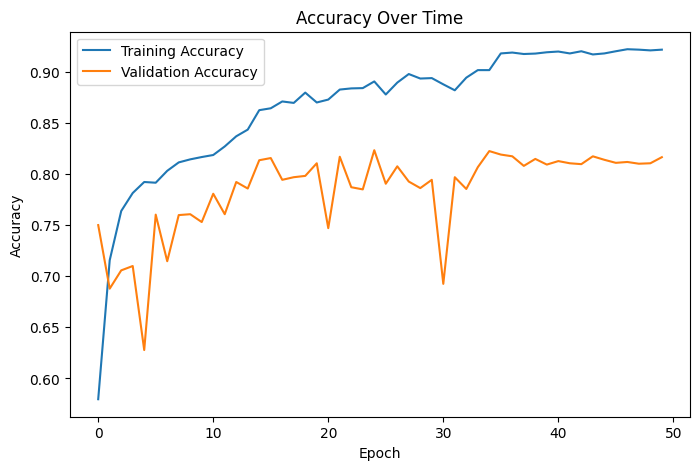}
  \label{fig112}
}
\hfill
\subfloat[Confusion matrix of 2D-Net with CNN-GRU feature extraction (EO-MA)]{
  \includegraphics[width=0.32\textwidth]{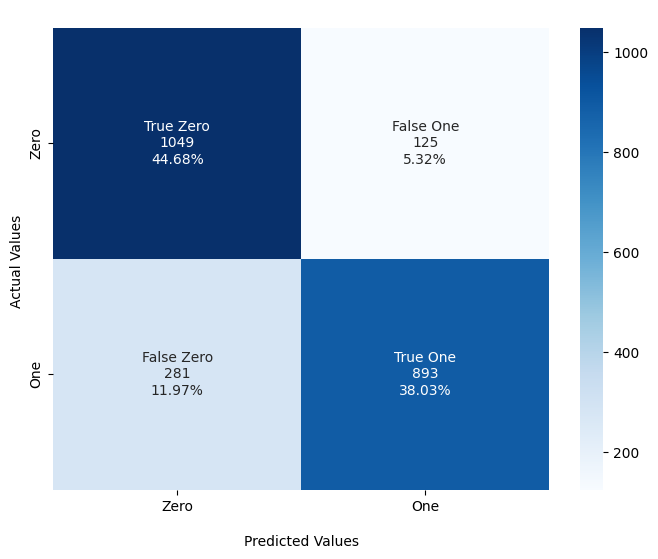}
  \label{fig113}
}
\hfill
\subfloat[Confusion matrix of 2D-Net without CNN-GRU feature extraction (EO-MA)]{
  \includegraphics[width=0.32\textwidth]{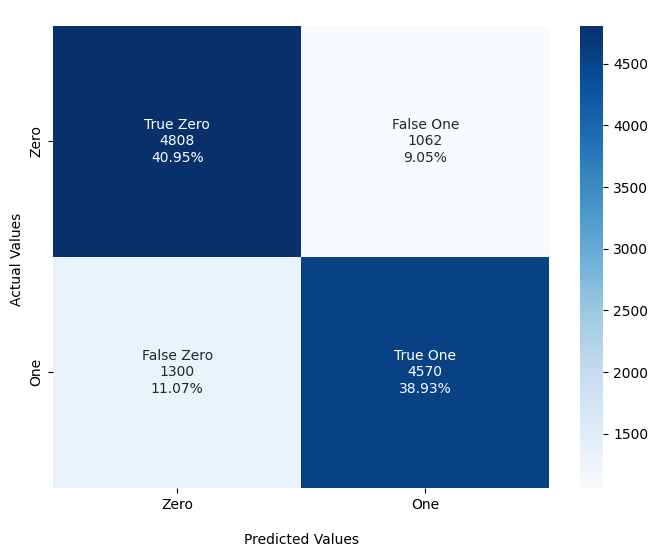}
  \label{fig55}
}
\caption{Performance of CNN-GRU and 2D-Net models (EO-MA)}
\label{fig1111}
\end{figure*}

The use of CNN-GRU features also improved the performance of the machine learning classifiers. XGBoost increased from 75.502\% to 82.368\%, SVM from 76.712\% to 82.410\%, and DT from 63.270\% to 82.538\%, as reported in Table \ref{tab111}. These results indicate a substantial improvement for the evaluated classifiers after CNN-GRU-based feature extraction. Table \ref{tab112} compares the classification accuracy obtained using different feature extraction methods. Among the evaluated approaches, CNN-GRU produced the highest accuracy for all four classifiers, with 82.368\% for XGBoost, 82.410\% for SVM, 82.538\% for DT, and 83.177\% for 2D-Net. Figure \ref{fig111} presents the corresponding accuracy comparison, while Fig. \ref{fig1111} shows the CNN-GRU accuracy across the 30 training epochs together with the confusion matrices for 2D-Net with and without CNN-GRU feature extraction.

\begin{table*}[!t]
\centering
\caption{Evaluation of DL and ML models with and without CNN-GRU feature extraction (EO-MA)}
\label{tab111}


\setlength{\tabcolsep}{6pt}
\renewcommand{\arraystretch}{1.15}

\begin{tabular}{lccccc ccccc}
\toprule
\multirow{2}{*}{\textbf{Model}} &
\multicolumn{5}{c}{\textbf{Without CNN-GRU Feature Extraction (\%)}} &
\multicolumn{5}{c}{\textbf{With CNN-GRU Feature Extraction (\%)}} \\
\cmidrule(lr){2-6}\cmidrule(lr){7-11}
& \textbf{Acc.} & \textbf{Prec.} & \textbf{Rec.} & \textbf{F1} & \textbf{Kappa}
& \textbf{Acc.} & \textbf{Prec.} & \textbf{Rec.} & \textbf{F1} & \textbf{Kappa} \\
\midrule
ANN      & 76.158 & 83.978 & 64.650 & 73.058 & 0.5231 & 81.942 & 88.109 & 73.850 & 80.352 & 0.6388 \\
1D CNN   & 79.855 & 84.062 & 73.679 & 78.529 & 0.5971 & 82.410 & 88.012 & 75.042 & 81.011 & 0.6482 \\
LSTM     & 78.705 & 88.243 & 66.235 & 75.671 & 0.5741 & 81.814 & 87.090 & 74.701 & 80.421 & 0.6362 \\
XGBoost  & 75.502 & 75.917 & 74.701 & 75.304 & 0.5100 & 82.368 & 86.679 & 76.490 & 81.267 & 0.6473 \\
SVM      & 76.712 & 79.275 & 72.333 & 75.645 & 0.5342 & 82.410 & 87.487 & 75.638 & 81.132 & 0.6482 \\
DT       & 63.270 & 62.749 & 65.315 & 64.006 & 0.2654 & 82.538 & 85.501 & 78.364 & 81.777 & 0.6507 \\
\textbf{2D-Net}   & \textbf{79.880} & \textbf{81.143} & \textbf{77.853} & \textbf{79.464} & \textbf{0.5976} &
\textcolor{blue}{\textbf{83.177}} & \textcolor{blue}{\textbf{88.298}} & \textcolor{blue}{\textbf{76.490}} & \textcolor{blue}{\textbf{81.971}} & \textcolor{blue}{\textbf{0.6635}} \\
\bottomrule
\end{tabular}
\end{table*}

\begin{table}[!t]
\centering
\caption{Accuracy score (\%) comparison of feature extraction methods. (EO-MA)}
\label{tab112}
\footnotesize

\begin{tabular}{lcccc}
\hline
\textbf{Method} & \textbf{XGBoost} & \textbf{SVM} & \textbf{DT} & \textbf{2D-Net} \\
\hline
AR & 58.560 & 54.063 & 58.892 & 50.017 \\
PCA & 76.013 & 78.211 & 69.131 & 74.914 \\
NMF & 57.333 & 55.579 & 51.950 & 55.783 \\
\textbf{CNN-GRU} & \textbf{82.368} & \textbf{82.410} & \textbf{82.538} & \textcolor{blue}{\textbf{83.177}} \\
\hline
\end{tabular}
\end{table}

\subsubsection{Resting-Memory (EO-ME) states classification}

The EO-ME task evaluates the distinction between the eyes-open resting state and the memory condition. Among the evaluated models, 2D-Net combined with CNN-GRU feature extraction attained the peak accuracy of 76.107\%, with the CNN-GRU features providing the input representation. Among the machine learning baselines, SVM achieved the highest accuracy at 75.553\% and also obtained the highest kappa value within that group.

\begin{figure}[!t]
\centering
\includegraphics[width=8.6cm]{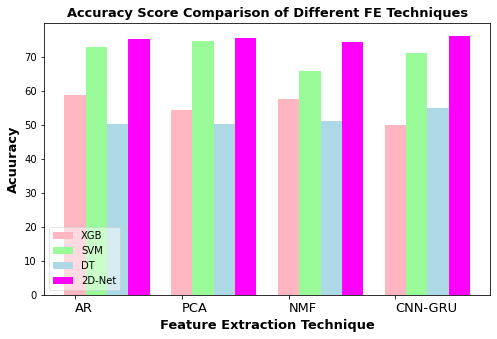}
\caption{Accuracy score (\%) comparison of different feature extraction techniques (EO-ME)}
\label{fig211}
\end{figure}

As in the EO-MA experiment, the use of CNN-GRU-based features generally improved classification performance. Table \ref{tab211} summarizes the results of the evaluated deep learning and machine learning models with and without CNN-GRU feature extraction. Table \ref{tab212} presents the accuracy obtained with the different feature extraction methods. CNN-GRU produced the highest accuracy for each of the evaluated classifiers, with 2D-Net achieving the overall best result. Figure \ref{fig211} compares the classification accuracy across the feature extraction methods, while Fig. \ref{fig2222} shows the CNN-GRU accuracy over the 30 training epochs together with the corresponding confusion matrices.

\begin{table*}[!t]
\centering
\caption{Evaluation of DL and ML models with and without CNN-GRU feature extraction (EO-ME)}
\label{tab211}

\setlength{\tabcolsep}{6pt}
\renewcommand{\arraystretch}{1.15}

\begin{tabular}{lccccc ccccc}
\toprule
\multirow{2}{*}{\textbf{Model}} &
\multicolumn{5}{c}{\textbf{Without CNN-GRU Feature Extraction (\%)}} &
\multicolumn{5}{c}{\textbf{With CNN-GRU Feature Extraction (\%)}} \\
\cmidrule(lr){2-6}\cmidrule(lr){7-11}
& \textbf{Acc.} & \textbf{Prec.} & \textbf{Rec.} & \textbf{F1} & \textbf{Kappa}
& \textbf{Acc.} & \textbf{Prec.} & \textbf{Rec.} & \textbf{F1} & \textbf{Kappa} \\
\midrule
ANN     & 73.356 & 75.965 & 68.330 & 71.946 & 0.4671 & 74.063 & 81.284 & 62.521 & 70.678 & 0.4812 \\
1D CNN  & 75.340 & 78.199 & 70.272 & 74.024 & 0.5068 & 74.659 & 80.250 & 65.417 & 72.078 & 0.4931 \\
LSTM    & 75.766 & 78.782 & 70.528 & 74.427 & 0.5153 & 74.488 & 79.792 & 65.587 & 71.996 & 0.4897 \\
XGBoost & 73.160 & 74.508 & 70.408 & 72.400 & 0.4632 & 75.170 & 78.999 & 68.569 & 73.415 & 0.5034 \\
SVM     & 73.364 & 74.278 & 71.482 & 72.853 & 0.4672 & 75.553 & 82.822 & 64.480 & 72.509 & 0.5110 \\
DT      & 64.787 & 65.764 & 61.686 & 63.660 & 0.2957 & 74.361 & 77.713 & 68.313 & 72.710 & 0.4872 \\
\textbf{2D-Net}
        & \textbf{74.480} & \textbf{74.673} & \textbf{74.088} & \textbf{74.380} & \textbf{0.4896}
        & \textcolor{blue}{\textbf{76.107}} & \textcolor{blue}{\textbf{81.893}} & \textcolor{blue}{\textbf{67.036}} & \textcolor{blue}{\textbf{73.723}} & \textcolor{blue}{\textbf{0.5221}} \\
\bottomrule
\end{tabular}
\end{table*}

\begin{table}[!t]
\centering
\caption{Accuracy score (\%) comparison of feature extraction methods (EO-ME)}
\label{tab212}
\footnotesize

\begin{tabular}{lcccc}
\hline
\textbf{Method} & \textbf{XGBoost} & \textbf{SVM} & \textbf{DT} & \textbf{2D-Net} \\
\hline
AR        & 58.816 & 54.190 & 57.683 & 50.000 \\
PCA       & 72.751 & 74.727 & 65.851 & 70.970 \\
NMF       & 50.187 & 50.195 & 51.039 & 55.008 \\
\textbf{CNN-GRU} &
\textbf{75.170} & \textbf{75.553} & \textbf{74.361} &
\textcolor{blue}{\textbf{76.107}} \\
\hline
\end{tabular}
\end{table}


\begin{figure}[!t]
\centering
\includegraphics[width=8.6cm]{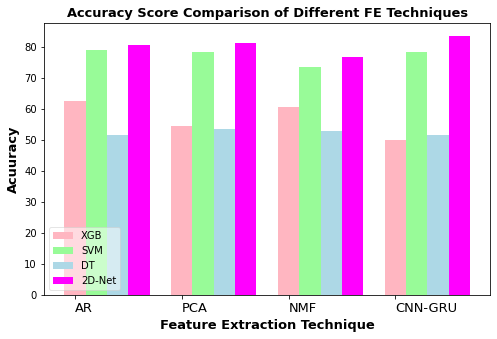}
\caption{Accuracy score (\%) comparison of different feature extraction techniques (EO-MU)}
\label{fig311}
\end{figure}

\begin{figure*}[!t]
\centering
\subfloat[GRU-CNN training vs validation accuracy (\%) (EO-ME)]{
  \includegraphics[width=0.32\textwidth]{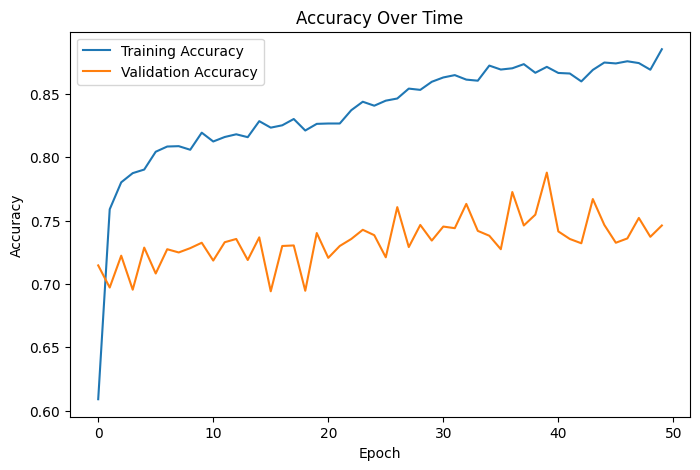}
  \label{fig212}
}
\hfill
\subfloat[Confusion matrix of 2D-Net with CNN-GRU feature extraction (EO-ME)]{
  \includegraphics[width=0.32\textwidth]{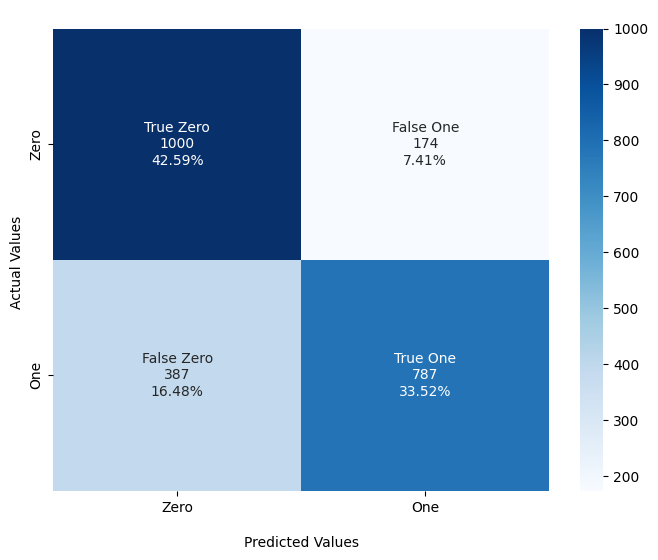}
  \label{fig213}
}
\hfill
\subfloat[Confusion matrix of 2D-Net without CNN-GRU feature extraction (EO-ME)]{
  \includegraphics[width=0.32\textwidth]{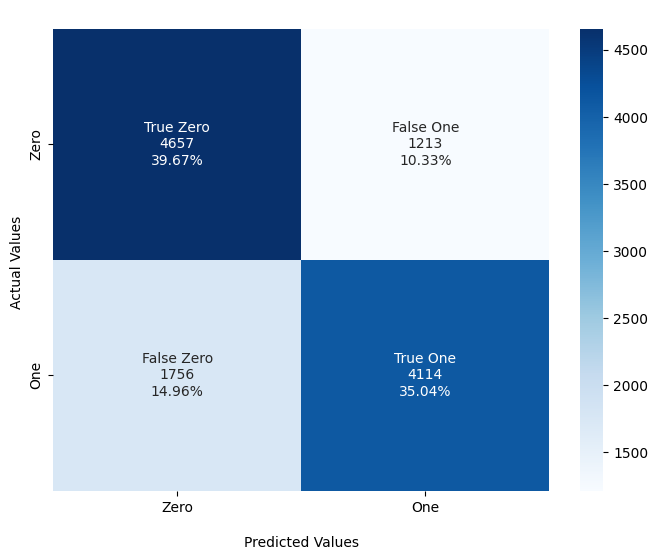}
  \label{fig214}
}
\caption{Performance of CNN-GRU and 2D-Net models (EO-ME)}
\label{fig2222}
\end{figure*}

\begin{table}[!t]
\centering
\caption{Accuracy score (\%) comparison of feature extraction methods (EO-MU)}
\label{tab312}
\footnotesize

\begin{tabular}{lcccc}
\hline
\textbf{Method} & \textbf{XGBoost} & \textbf{SVM} & \textbf{DT} & \textbf{2D-Net} \\
\hline
AR        & 62.325 & 54.437 & 60.383 & 50.000 \\
PCA       & 78.892 & 78.313 & 73.466 & 78.381 \\
NMF       & 51.388 & 53.569 & 52.904 & 51.575 \\
\textbf{CNN-GRU} &
\textbf{80.621} & \textbf{81.218} & \textbf{76.618} &
\textcolor{blue}{\textbf{83.432}} \\
\hline
\end{tabular}
\end{table}

\begin{figure*}[!t]
\centering
\subfloat[GRU-CNN training vs validation accuracy (\%) (EO-MU)]{
  \includegraphics[width=0.32\textwidth]{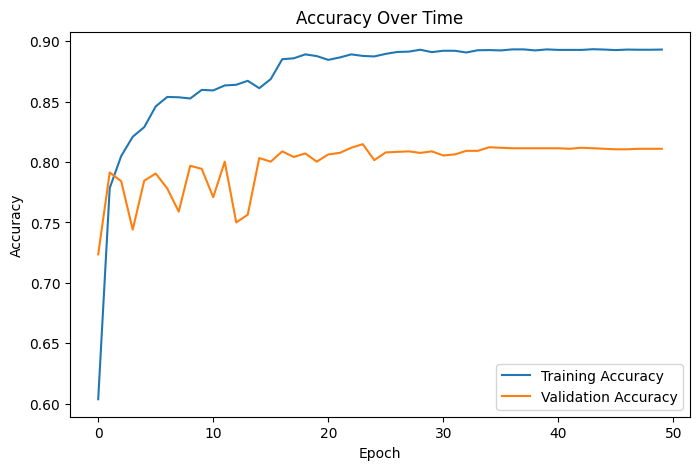}
  \label{fig312}
}
\hfill
\subfloat[Confusion matrix of 2D-Net with CNN-GRU feature extraction (EO-MU)]{
  \includegraphics[width=0.32\textwidth]{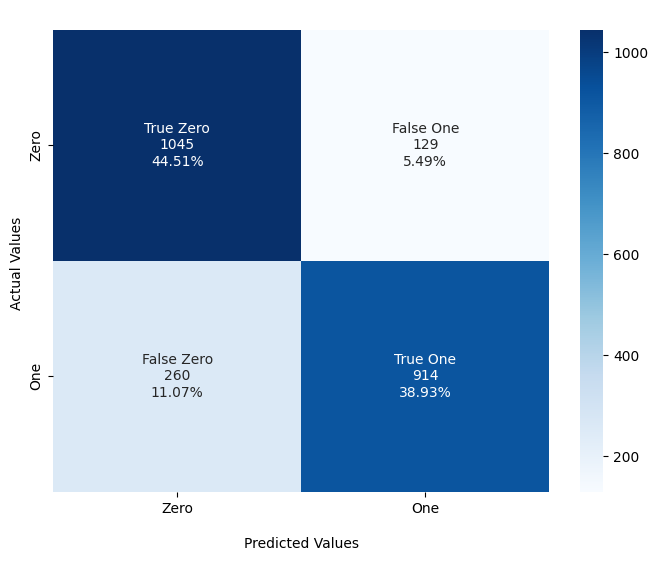}
  \label{fig313}
}
\hfill
\subfloat[Confusion matrix of 2D-Net without CNN-GRU feature extraction (EO-MU)]{
  \includegraphics[width=0.32\textwidth]{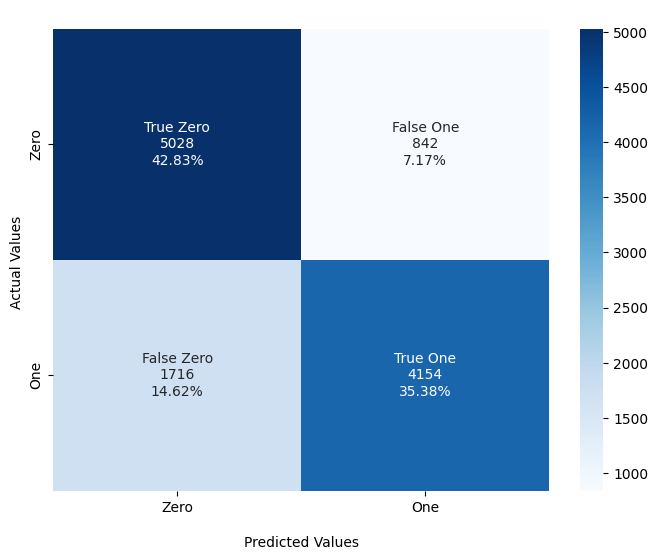}
  \label{fig314}
}
\caption{Performance of CNN-GRU and 2D-Net models (EO-MU)}
\label{fig333}
\end{figure*}

\begin{table*}[!t]
\centering
\caption{Evaluation of DL and ML models with and without CNN-GRU feature extraction (EO-MU)}
\label{tab311}

\setlength{\tabcolsep}{6pt}
\renewcommand{\arraystretch}{1.15}

\begin{tabular}{lccccc ccccc}
\toprule
\multirow{2}{*}{\textbf{Model}} &
\multicolumn{5}{c}{\textbf{Without CNN-GRU Feature Extraction (\%)}} &
\multicolumn{5}{c}{\textbf{With CNN-GRU Feature Extraction (\%)}} \\
\cmidrule(lr){2-6}\cmidrule(lr){7-11}
& \textbf{Acc.} & \textbf{Prec.} & \textbf{Rec.} & \textbf{F1} & \textbf{Kappa}
& \textbf{Acc.} & \textbf{Prec.} & \textbf{Rec.} & \textbf{F1} & \textbf{Kappa} \\
\midrule
ANN     & 80.775 & 83.385 & 76.865 & 79.992 & 0.6155 & 83.177 & 87.778 & 77.086 & 82.086 & 0.6635 \\
1D CNN  & 78.918 & 78.553 & 79.557 & 79.052 & 0.5783 & 83.049 & 86.397 & 78.449 & 82.232 & 0.6609 \\
LSTM    & 80.741 & 83.238 & 76.984 & 79.989 & 0.6148 & 82.751 & 87.149 & 76.831 & 81.665 & 0.6550 \\
XGBoost & 76.192 & 76.494 & 75.621 & 76.055 & 0.5238 & 80.621 & 85.073 & 74.276 & 79.308 & 0.6124 \\
SVM     & 77.913 & 79.495 & 75.230 & 77.304 & 0.5582 & 81.218 & 87.436 & 72.913 & 79.517 & 0.6243 \\
DT      & 69.165 & 70.864 & 65.093 & 67.856 & 0.3833 & 76.618 & 78.908 & 72.657 & 75.654 & 0.5323 \\
\textbf{2D-Net}
        & \textbf{78.211} & \textbf{83.146} & \textbf{70.766} & \textbf{76.458} & \textbf{0.5642}
        & \textcolor{blue}{\textbf{83.432}} & \textcolor{blue}{\textbf{87.631}} & \textcolor{blue}{\textbf{77.853}} & \textcolor{blue}{\textbf{82.453}} & \textcolor{blue}{\textbf{0.6686}} \\
\bottomrule
\end{tabular}
\end{table*}

\subsubsection{Resting-Music (EO-MU) states classification}

The EO-MU experiment compares the eyes-open resting condition with the music-related mental task using the baseline models and the proposed 2D-Net. The highest performance was obtained with 2D-Net combined with CNN-GRU feature extraction, achieving 83.432\% accuracy and a Cohen's kappa of 0.6686.

CNN-GRU feature extraction also improved the performance of the machine learning classifiers. The largest increase was observed for the Decision Tree, whose accuracy rose from 69.165\% to 76.618\%. The results for the evaluated deep learning and machine learning models are presented in Table \ref{tab311}, including the metrics introduced earlier. Table \ref{tab312} compares the accuracy obtained using the different feature extraction methods. CNN-GRU achieved the highest accuracy for all evaluated classifiers, including XGBoost, SVM, DT, and 2D-Net. Figure \ref{fig311} presents the accuracy comparison among the feature extraction methods, while Fig. \ref{fig333} shows the CNN-GRU accuracy over the 30 training epochs together with the confusion matrices for 2D-Net with and without CNN-GRU feature extraction.



\section{Conclusion}
\label{Sec:6}

This study introduced a signal-processing and deep learning framework for the binary classification of resting and cognitive EEG states. The proposed approach combined CNN-GRU-based feature extraction with the 2D-Net classifier to distinguish the eyes-open resting condition from mathematical, memory, and music-related cognitive tasks. Across the three classification tasks, the proposed approach achieved accuracies of 83.177\% for EO-MA, 76.107\% for EO-ME, and 83.432\% for EO-MU, outperforming the evaluated baseline models. The results also showed that CNN-GRU-based feature extraction enhanced the classification efficacy of the assessed machine learning and deep learning models. The frequency-domain analysis additionally uncovered variations in the distribution of EEG power among the considered conditions, with prominent alpha-band activity observed in the eyes-closed and cognitive states in the analyzed data. Overall, the findings support the application of combined signal processing and deep learning techniques for distinguishing resting and cognitive states from EEG recordings.

\section{Limitations and scope for further research}
\label{Sec:7}

The dataset contains recordings from 60 participants; however, only 40 participants were included in the present experiments  to alleviate the computational load of model training. Furthermore, solely the initial trial was examined, whilst the subsequent second and third trials were omitted from the experiments. The second trial was excluded to reduce the possibility of bias related to repeated exposure to the tasks, whereas the third trial was omitted to keep the analysis within the selected experimental scope. Another limitation is that the study considers binary classification only, which does not address more complex settings that require simultaneous discrimination among multiple cognitive states. Future studies can extend the analysis by incorporating all three trials and examining the test-retest characteristics of the EEG recordings. The questionnaire information available with the dataset can also be incorporated to investigate relationships between reported subjective or emotional states and corresponding EEG patterns. Such an extension could provide a broader view of the relationship between self-reported experiences and measured brain activity.

\EOD


\begin{thebibliography}{00}

\bibitem{b1} R. F. Ahmad, A. S. Malik, H. U. Amin, N. Kamel, and F. Reza, ``Classification of cognitive and resting states of the brain using EEG features,'' in \emph{Proc. IEEE Int. Symp. Med. Meas. Appl. (MeMeA)}, 2016, pp. 1--5.

\bibitem{b2} S. Chawla, R. Ranjan, and Y. Narayan, ``A systematic review of EEG signals classification using machine learning and deep learning approach,'' in \emph{Proc. IEEE Uttar Pradesh Section Int. Conf. Elect., Electron. Comput. Eng. (UPCON)}, 2023, pp. 1228--1232.

\bibitem{b3} S. C. Yener, A. Uygur, and H. H. Kuntman, ``Ultra low-voltage ultra low-power memristor based band-pass filter design and its application to EEG signal processing,'' \emph{Analog Integr. Circuits Signal Process.}, vol. 89, pp. 719--726, 2016.

\bibitem{b4} J. Baranowski and P. Pi{\k{a}}tek, ``Fractional band-pass filters: Design, implementation and application to EEG signal processing,'' \emph{J. Circuits, Syst. Comput.}, vol. 26, no. 11, Art. no. 1750170, 2017.

\bibitem{b5} S. S. Daud and R. Sudirman, ``Butterworth bandpass and stationary wavelet transform filter comparison for electroencephalography signal,'' in \emph{Proc. Int. Conf. Intell. Syst., Model. Simul.}, 2015, pp. 123--126.

\bibitem{b6} X. Hu and Q. Yuan, ``Epileptic EEG identification based on deep Bi-LSTM network,'' in \emph{Proc. IEEE Int. Conf. Adv. Infocomm Technol. (ICAIT)}, 2019, pp. 63--66.

\bibitem{b7} S. Abenna, M. Nahid, and H. Bouyghf, ``Sleep stages detection based BCI: A novel single-channel EEG classification based on optimized bandpass filter,'' in \emph{Adv. Technol. Humanity (Proc. ICATH'2021)}. Springer, 2022, pp. 96--105.

\bibitem{b8} V. Vimala, K. Ramar, and M. Ettappan, ``An intelligent sleep apnea classification system based on EEG signals,'' \emph{J. Med. Syst.}, vol. 43, no. 2, Art. no. 36, 2019.

\bibitem{b9} A. Subasi and M. I. Gursoy, ``EEG signal classification using PCA, ICA, LDA and support vector machines,'' \emph{Expert Syst. Appl.}, vol. 37, no. 12, pp. 8659--8666, 2010.

\bibitem{b10} I. Winkler, S. Haufe, and M. Tangermann, ``Automatic classification of artifactual ICA-components for artifact removal in EEG signals,'' \emph{Behav. Brain Funct.}, vol. 7, pp. 1--15, 2011.

\bibitem{b11} L. Albera \emph{et al.}, ``ICA-based EEG denoising: A comparative analysis of fifteen methods,'' \emph{Bull. Polish Acad. Sci.: Tech. Sci.}, vol. 60, no. 3, pp. 407--418, 2012.

\bibitem{b12} M. Murugappan and S. Murugappan, ``Human emotion recognition through short time electroencephalogram (EEG) signals using fast Fourier transform (FFT),'' in \emph{Proc. IEEE Int. Colloq. Signal Process. Appl.}, 2013, pp. 289--294.

\bibitem{b13} K. Polat and S. G{\"u}ne{\c{s}}, ``Classification of epileptiform EEG using a hybrid system based on decision tree classifier and fast Fourier transform,'' \emph{Appl. Math. Comput.}, vol. 187, no. 2, pp. 1017--1026, 2007.

\bibitem{b14} S. K. Hadjidimitriou and L. J. Hadjileontiadis, ``Toward an EEG-based recognition of music liking using time-frequency analysis,'' \emph{IEEE Trans. Biomed. Eng.}, vol. 59, no. 12, pp. 3498--3510, 2012.

\bibitem{b15} Z. Wang \emph{et al.}, ``Short time Fourier transformation and deep neural networks for motor imagery brain computer interface recognition,'' \emph{Concurrency Comput.: Pract. Exper.}, vol. 30, no. 23, Art. no. e4413, 2018.

\bibitem{b16} F. Barneih \emph{et al.}, ``Artificial neural network model using short-term Fourier transform for epilepsy seizure detection,'' in \emph{Proc. Adv. Sci. Eng. Technol. Int. Conf. (ASET)}, 2022, pp. 1--5.

\bibitem{b17} A. A. Vergani, ``Hans Berger (1873--1941): The German psychiatrist who recorded the first electrical brain signal in humans 100 years ago,'' \emph{Adv. Physiol. Educ.}, vol. 48, no. 4, pp. 878--881, 2024, doi: 10.1152/advan.00119.2024.

\bibitem{b18} A. Hyv\"arinen, ``Fast and robust fixed-point algorithms for independent component analysis,'' \emph{IEEE Trans. Neural Netw.}, vol. 10, no. 3, pp. 626--634, 1999.

\bibitem{b19} L. Cohen, \emph{Time-Frequency Analysis}. Upper Saddle River, NJ, USA: Prentice Hall, 1995.

\bibitem{b20} S. Siuly, Y. Li, and Y. Zhang, ``EEG signal analysis and classification,'' \emph{IEEE Trans. Neural Syst. Rehabil. Eng.}, 2016.

\bibitem{b21} V. Perez, A. Duque, V. Hidalgo, and A. Salvador, ``EEG frequency bands in subjective cognitive decline: A systematic review of resting state studies,'' \emph{Biol. Psychol.}, vol. 191, Art. no. 108823, 2024.

\bibitem{b22} K. Cho \emph{et al.}, ``Learning phrase representations using RNN encoder-decoder for statistical machine translation,'' \emph{Proc. 2014 Conf. Empirical Methods in Natural Language Processing (EMNLP)}, 2014, pp. 1724--1734.

\bibitem{b23} S. Kiranyaz, O. Avci, O. Abdeljaber, T. Ince, M. Gabbouj, and D. J. Inman, ``1D convolutional neural networks and applications: A survey,'' \emph{Mech. Syst. Signal Process.}, vol. 151, Art. no. 107398, 2021.

\bibitem{b24} A. Craik, Y. He, and J. L. Contreras-Vidal, ``Deep learning for electroencephalogram (EEG) classification tasks: A review,'' \emph{J. Neural Eng.}, vol. 16, no. 3, Art. no. 031001, 2019.

\bibitem{b25} J. M. Anderson, R. J. M. Peters, and A. C. C. Coolen, ``EEG-based classification of cognitive states,'' \emph{Biol. Cybern.}, 1996.

\bibitem{b26} Y. Wang, W. Duan, D. Dong, L. Ding, and X. Lei, ``A test-retest resting and cognitive state EEG dataset during multiple subject-driven states,'' \emph{Sci. Data}, vol. 9, Art. no. 566, 2022.

\bibitem{b27} Z. Bai, R. Yang, and Y. Liang, ``Mental task classification using electroencephalogram signal,'' \emph{arXiv preprint} arXiv:1910.03023, 2019.

\bibitem{b28} F. Siddiqui \emph{et al.}, ``Deep neural network for EEG signal-based subject-independent imaginary mental task classification,'' \emph{Diagnostics}, vol. 13, no. 4, Art. no. 640, 2023.

\bibitem{b29} P. L. Lee, S. H. Chen, T. C. Chang, W. K. Lee, H. T. Hsu, and H. H. Chang, ``Continual learning of a transformer-based deep learning classifier using an initial model from action observation EEG data to online motor imagery classification,'' \emph{Bioengineering}, vol. 10, no. 2, Art. no. 186, 2023.

\bibitem{b30} A. Qayyum, M. K. A. A. Khan, M. Mazher, and M. Suresh, ``Classification of EEG learning and resting states using 1D-convolutional neural network for cognitive load assessment,'' in \emph{Proc. IEEE Student Conf. Res. Develop. (SCOReD)}, 2018, pp. 1--5.

\bibitem{b31} M. Mazher, I. Faye, A. Qayyum, and A. S. Malik, ``Classification of resting and cognitive states using EEG-based feature extraction and connectivity approach,'' in \emph{Proc. IEEE-EMBS Conf. Biomed. Eng. Sci. (IECBES)}, 2018, pp. 184--188.

\bibitem{b32} N. P. Tigga and S. Garg, ``Efficacy of novel attention-based gated recurrent units transformer for depression detection using electroencephalogram signals,'' \emph{Health Inf. Sci. Syst.}, vol. 11, no. 1, pp. 1--17, 2023.

\bibitem{b33} N.-Y. Liang, P. Saratchandran, G.-B. Huang, and N. Sundararajan, ``Classification of mental tasks from EEG signals using extreme learning machine,'' \emph{Int. J. Neural Syst.}, vol. 16, no. 1, pp. 29--38, 2006.

\bibitem{b34} Z. Li and M. Shen, ``Classification of mental task EEG signals using wavelet packet entropy and SVM,'' in \emph{Proc. Int. Conf. Electron. Meas. Instrum.}, 2007, pp. 3--906.

\bibitem{b35} N. Panwar, V. Pandey, and P. P. Roy, ``EEG-CogNet: A deep learning framework for cognitive state assessment using EEG brain connectivity,'' \emph{Biomed. Signal Process. Control}, vol. 98, Art. no. 106770, 2024.

\bibitem{b36} Y. Wang, M. Han, Y. Peng, R. Zhao, D. Fan, X. Meng, H. Xu, H. Niu, J. Cheng, and T. Liu, ``LGNet: Learning local--global EEG representations for cognitive workload classification in simulated flights,'' \emph{Biomed. Signal Process. Control}, vol. 92, Art. no. 106046, 2024.

\bibitem{b37} S. J. Sharma and R. Gupta, ``Deep learning-based EEG-based mental workload detection with discrete wavelet transform and Welch's power spectral density,'' \emph{Procedia Comput. Sci.}, vol. 260, pp. 134--141, 2025.

\bibitem{b38} P. D. Welch, ``The use of fast Fourier transform for the estimation of power spectra: A method based on time averaging over short, modified periodograms,'' \emph{IEEE Trans. Audio Electroacoust.}, vol. 15, no. 2, pp. 70--73, 1967, doi: 10.1109/TAU.1967.1161901.

\bibitem{b39} J. Cohen, ``A coefficient of agreement for nominal scales,'' \emph{Educ. Psychol. Meas.}, vol. 20, no. 1, pp. 37--46, 1960.

\bibitem{b40} OpenNeuro, ``Test-retest resting and cognitive state EEG dataset during multiple subject-driven states (ds004148), ver. 1.0.1.'' Accessed on: Dec. 5, 2022. [Online]. Available: \url{https://openneuro.org/datasets/ds004148/versions/1.0.1//}

\end{thebibliography}
\end{document}